%% file: final_draft.tex
\documentclass[10pt,journal,compsoc]{IEEEtran}
\ifCLASSOPTIONcompsoc
  \usepackage[nocompress]{cite}
\else
  \usepackage{cite}
\fi
\ifCLASSINFOpdf
\else
\fi
\usepackage{amsmath}
\usepackage{amssymb}
\usepackage{algorithm}
\usepackage{algorithmic}
\usepackage{subfigure}
\usepackage{graphicx}
\usepackage{tabularx}
\usepackage{color}
\usepackage{caption}
\usepackage{ragged2e}
\usepackage{enumitem}
\input{macro3}

\pdfoutput=1
\usepackage{etoolbox,lipsum}

\newcommand{\etal}{{\it{et al. }}}

\begin{document}
%
\title{Superpixel Soup: Monocular Dense 3D Reconstruction of a Complex Dynamic Scene}%
%
%
%

\author{Suryansh Kumar,~\IEEEmembership{Member, ~IEEE,}
        Yuchao Dai,~\IEEEmembership{Member,~IEEE,}
        Hongdong Li,~\IEEEmembership{Senior Member,~IEEE}
\IEEEcompsocitemizethanks{\IEEEcompsocthanksitem Suryansh Kumar is with ETH Z\"urich and Australian National University. \protect
E-mail: sukumar@vision.ee.ethz.ch, suryansh.kumar@anu.edu.au.
\IEEEcompsocthanksitem Yuchao Dai is with Northwestern Polytechnical University. E-mail: daiyuchao@gmail.com.
\IEEEcompsocthanksitem Hongdong Li is with Australian National University and ARC Centre of Excellent for Robotic Vision. E-mail: hongdong.li@anu.edu.au.}
}

\IEEEtitleabstractindextext{
\justify
\begin{abstract}
This work addresses the task of dense 3D reconstruction of a complex dynamic scene from images. The prevailing idea to solve this task is composed of a sequence of steps and is dependent on the success of several pipelines in its execution \cite{ranftl2016dense}. To overcome such limitations with the existing algorithm, we propose a unified approach to solve this problem. We assume that a dynamic scene can be approximated by numerous piecewise planar surfaces, where each planar surface enjoys its own rigid motion, and the global change in the scene between two frames is as-rigid-as-possible (ARAP). Consequently, our model of a dynamic scene reduces to a \emph{soup} of planar structures and rigid motion of these local planar structures. Using planar over-segmentation of the scene, we reduce this task to solving a ``3D jigsaw puzzle'' problem. Hence, the task boils down to correctly assemble each rigid piece to construct a 3D shape that complies with the geometry of the scene under the ARAP assumption. Further, we show that our approach provides an effective solution to the inherent scale-ambiguity in structure-from-motion under perspective projection. We provide extensive experimental results and evaluation on several benchmark datasets. Quantitative comparison with competing approaches shows state-of-the-art performance.

\end{abstract}

}

\maketitle

\IEEEdisplaynontitleabstractindextext

%
\IEEEpeerreviewmaketitle

\ifCLASSOPTIONcompsoc
\IEEEraisesectionheading{\section{Introduction}\label{sec:introduction}}
\else
\section{Introduction}
\label{sec:introduction}
\fi

\IEEEPARstart{T}{he} task of reconstructing 3D geometry of the scene from images ---popularly known as structure-from-motion (SfM), is a fundamental problem in computer vision. An initial 
introduction and working solution to this problem can be found as early as 1970's and 1980's \cite{ullman1979interpretation} \cite{grimson1981images} \cite{longuet1981computer}, which Blake \etal discussed comprehensively in their 
seminal work \cite{blake1987visual}. While this field of study in the past was largely dominated by sparse feature based reconstruction of a rigid scene \cite{hartley1997triangulation} \cite{hartley1997defense} \cite{hartley2003multiple} \cite{tomasi1993pictures} \cite{tomasi1992shape} and a non-rigid object \cite{bregler2000recovering} \cite{dai2014simple} \cite{lee2013procrustean} \cite{kumar2016multi} \cite{kumar2017spatio}, in recent years, with the surge 
in computational resources, dense 3D reconstruction of the scene have been introduced and successfully demonstrated \cite{newcombe2015dynamicfusion} \cite{newcombe2011dtam} \cite{ranftl2016dense}.

A dense solution to this inverse problem is essential due to its increasing demand in many real-world applications --from animation and entertainment industry to robotics industry (VSLAM). In particular, with the proliferation of \emph{monocular} camera in almost all modern mobile devices has elevated the demand for sophisticated dense reconstruction algorithm. When the scene is static and the camera is moving, 3D reconstruction of such scenes from images can be achieved by using conventional rigid structure from motion techniques \cite{hartley2003multiple} \cite{agarwal2011building} \cite{schoenberger2016sfm} \cite{schoenberger2016mvs}. In contrast, to model arbitrary dynamic scene can be very challenging. When the camera is moving and the scene is static under such settings, the elegant geometrical constraint can help explain the camera's \cite{hartley1997defense} \cite{govindu2001combining}, which are later used to realize the dense 3D reconstruction of the scene \cite{schoenberger2016sfm} \cite{schoenberger2016mvs} \cite{newcombe2011dtam} \cite{triggs1999bundle}. However, such geometrical constraint may fail when multiple rigidly moving objects are observed by a moving camera. Although each of the individual rigid objects can be reconstructed up to an arbitrary scale (assuming motion segmentation is provided), the reconstruction of the whole dynamic scene is generally impossible, simply because the relative scales among all the moving shapes cannot be determined in a globally consistent way. Furthermore, since all the estimated motions are relative to each other, one cannot distinguish camera motion from the object motion. Therefore, prior information about the objects, or the scene, and their relation to the frame of reference are used to fix the placement of these objects relative to each other.

\begin{figure}
\centering
\includegraphics[width=0.5\textwidth, height = 0.15\textheight] {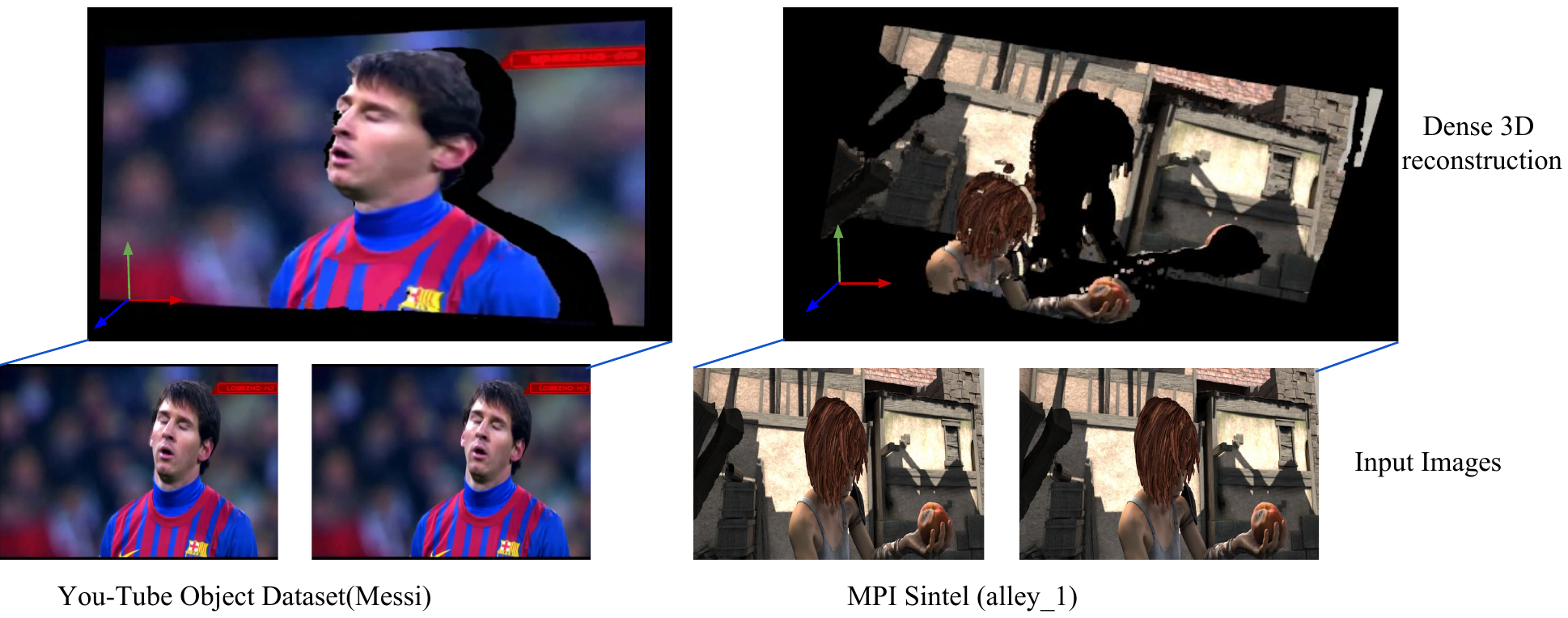}
\caption{\scriptsize Dense 3D reconstruction of a complex dynamic scene, where both the camera and the objects are moving with respect to each other. The top left shows a sample reconstruction on messi sequence from Youtube Object dataset \cite{prest2012learning}. The top right shows the reconstruction on alley\_1 sequence from the MPI Sintel dataset \cite{butler2012naturalistic}.}
\label{fig:firstSampleResult}
\end{figure}

Hence, from the above discussion, it can be argued that the solution to 3D reconstruction of a general dynamic scene is non-trivial. Nevertheless, it is an important problem to solve as many real-world applications need a reliable solution to this problem. For example, understanding of a traffic scene, a typical outdoor traffic scene consists of both multiple rigid motions of vehicles, and non-rigid motion of the pedestrians. To model such scenarios, it is important to have an algorithm that can provide dense 3D information from images.

Recently, Ranftl \etal \cite{ranftl2016dense} proposed a three-step approach to procure dense 3D reconstruction of a general dynamic scene using two consecutive perspective frames. Concretely, it performs object-level motion segmentation followed by per-object 3D reconstruction and finally solves for scale ambiguity. We know that in a general dynamic setting, the task of densely segmenting rigidly moving objects or part is not trivial.  Consequently, inferring motion models for deforming shapes becomes very challenging. Furthermore, the success of object-level segmentation builds upon the assumption of multiple rigid motions, fails to describe more general scenarios such as ``when the objects themselves are deforming''. Subsequently, 3D reconstruction algorithms dependent on motion segmentation of objects suffer.

Motivated by such limitations, we propose a unified approach that neither performs any object-level motion segmentation nor assumes any prior knowledge about the scene rigidity type and still able to recover scale consistent dense reconstruction of a complex dynamic scene. Our formulation instinctively encapsulates the solution to inherent scale ambiguity in perspective structure from motion which is a very challenging problem in general. We show that by using \emph{two prior assumptions} ---about the 3D scene and the deformation, we can effectively pin down the unknown relative scales, and obtain a globally consistent dense 3D reconstruction of a dynamic scene from its two perspective views. The two basic assumptions we used about the dynamic scene are:

\begin{enumerate}
  \item The dynamic scene can be approximated by a collection of \emph{piecewise planar surfaces} each having its own \emph{rigid motion}.
  \item The deformation of the scene between two frames is {\em locally-rigid} but {\em globally as-rigid-as-possible}.
  \end{enumerate}
  
\begin{figure}[t]
\begin{center}
\includegraphics[width=1.0\linewidth]{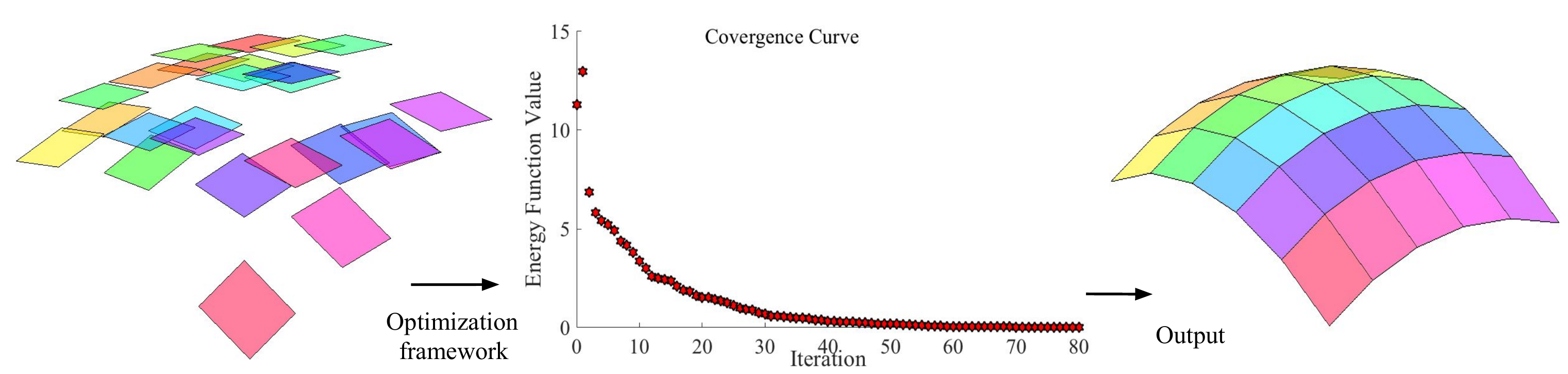}
\caption{\scriptsize  Reconstructing a 3D surface from a soup of un-scaled superpixels via solving a 3D Superpixel Jigsaw puzzle problem.\label{fig:concept}}
\end{center}
\end{figure}

  \begin{itemize}[leftmargin=*]
      \item \emph{Piece-wise planar model}: Our method models a dynamic scene as a collection of \emph{piece-wise planar} regions. Given two perspective images $\mathbf{I}$ (reference image), $\mathbf{I}'$ (next image) of a general dynamic scene, our method first over-segment the reference image into superpixels. This collection of superpixels are assumed approximation of the dynamic scene in the projective space. It can be argued that modeling dynamic scene per pixel can be more compelling, however, modeling of a scene using planar regions makes this problem computationally tractable for optimization or inference \cite{bleyer2011object, vogel20153d}.
  
      \item \emph{Locally-rigid and globally as-rigid-as-possible}: We implicitly assume that each local plane undergoes a \emph{rigid motion}. Suppose every individual superpixel corresponds to a small planar patch moving rigidly in 3D space and dense optical flow between frame is given, we can estimate its location in 3D using rigid reconstruction pipeline \cite{hartley2003multiple, vogel20113d}. Since the \emph{relative scale} of these patches is not determined correctly, they are floating in 3D space as a set of unorganized superpixel soup. Under the assumption that the change between the frame is not too arbitrary rather regular or smooth, the scene can be assumed to be changing as rigid as possible globally. Using this intuition, our method starts finding for each superpixel an appropriate scale, under which the entire set of superpixels can be assembled (glued) together coherently, forming a piece-wise smooth surface, {\em as if} playing the game of ``3D jigsaw puzzle".  Hence, we call our method the ``SuperPixel Soup'' algorithm (see Fig. \ref{fig:concept} for a conceptual visualization).
  \end{itemize}


In this paper, we show that our aforementioned assumptions can faithfully model most of the real-world dynamic scenarios. Furthermore, we encapsulate these assumptions in a simple optimization problem which are solved using a combination of continuous and discrete optimization algorithms \cite{benson2002interior, benson2014interior, kolmogorov2006convergent}. We demonstrate the benefit of our approach on available benchmark dataset such as KITTI \cite{geiger2013vision}, MPI Sintel \cite{butler2012naturalistic} and Virtual KITTI \cite{gaidon2016virtual}. The statistical comparison shows that our algorithm outperforms many available state-of-the-art methods by a significant margin.

\section{Related Works}
The solution to SfM has undergone prodigious development since its inception \cite{ullman1979interpretation}. Even after such a remarkable development in this field, the choice of algorithm depends on the complexity of the object motion and the environment. In this work, we utilize the idea of rigidity (locally) to solve dense reconstruction of a general dynamic scene. The concept of rigidity is not new in structure from motion problem \cite{ullman1979interpretation} \cite{longuet1987computer} and has been effectively applied as a global constraint to solve large scale reconstruction problem \cite{agarwal2011building}. The idea of global rigidity to solve structure and motion has also been 
exploited to solve reconstruction over multiple frames via a factorization approach \cite{tomasi1992shape}.

The literature on structure from motion and its treatment to different scenarios is very extensive. Consequently, for brevity, 
we only discuss the previous works that are of direct relevance to dynamic 3D reconstruction from \emph{monocular} images. The linear low-rank model has been used for dense non-rigid reconstruction. Kumar \etal \cite{kumar2018scalable, Kumar_2019_CVPR} and Garg \etal \cite{garg2013dense} solved the task with an orthographic camera model assuming feature matches across multiple frames is given as input. Fayad \etal \cite{fayad2010piecewise} recovered deformable surfaces with a quadratic approximation, again from multiple frames.  Taylor \etal \cite{taylor2010non} proposed a piecewise rigid solution using locally-rigid SfM to reconstruct a soup of rigid triangles.

While Taylor \etal \cite{taylor2010non} method is conceptually similar to ours, there are major differences:
\begin{enumerate}
\item We achieve \textit{two-view} dense reconstruction while \cite{taylor2010non} relies on multiple views (N $\geq$ 4).
\item We use \textit{perspective camera model} while they rely on an orthographic camera model.
\item We solve the scale-indeterminacy issue, which is an inherent ambiguity for 3D reconstruction under perspective projection, while Taylor \etal \cite{taylor2010non} method does not suffer from this, at the cost of being restricted to the orthographic camera model.
\end{enumerate}

Recently, Russel \etal \cite{russell2014video} and Ranftl \etal \cite{ranftl2016dense} used object-level segmentation for dense dynamic 3D reconstruction. In contrast, our method is free from object segmentation, hence circumvent the difficulty associated with motion segmentation in a dynamic setting.

The template-based approach is yet another method for deformable surface reconstruction. Yu \etal \cite{yu2015direct} proposed a direct approach to capture dense, detailed 3D geometry of generic, complex non-rigid meshes using a single RGB camera. While it works for generic surfaces, the requirement of template prevents its wider application to more general scenes.  Wang \etal \cite{wang2016template} introduced a template-free approach to reconstruct a poorly-textured, deformable surface. Nevertheless, its success is restricted to a single deforming surface rather than the entire dynamic scene. Varol \etal \cite{varol2009template} reconstructed deformable surfaces based on a piecewise reconstruction assuming overlapping patches to be consistent over the entire surface, but again limited to the reconstruction of a single deformable surface.

While the conceptual idea of our work appeared in 
ICCV 2017, this journal version provides \emph{(i)} in-depth realization of our overall optimization 
\emph{(ii)} Qualitative comparison with \cite{ranftl2016dense}, Video-PopUp \cite{russell2014video} 
as well as statistical comparison with deep-learning method \cite{zhou2017unsupervised}.
\emph{(iii)} Comprehensive ablation test showing the importance of each term in the overall optimization.
\emph{(iv)} Extensive performance analysis showing the performance with the variation in the number of superpixels, choice of k-nearest neighbor, choice of dense optical flow algorithm and change in the shape of the superpixel.
\emph{(v)} Detail discussion on the failure cases, choice of euclidean metric for nearest neighbor graph construction, and limitation of our work with possible direction for improvements.

\section{Motivation and Contribution}
The formulation proposed in this work is motivated by the following endeavor in dense structure from motion of a dynamic scene.

\subsection{Object level motion segmentation}
To solve dense reconstruction of an entire dynamic scene from perspective images, the first step that is practiced usually is: Perform object-level motion segmentation to infer distinct motion models of multiple rigidly moving object in the scene. As alluded before, dense segmentation of moving object in a dynamic scene in itself is a challenging task. Also, non-rigidly moving object themselves may compose of a union of distinct motion models. Therefore, object-level segmentation build upon the assumption of per object rigid motion will fail to describe a 
general dynamic scene. This motivates us to develop an algorithm that can recover a dense-detailed 3D model of a 
complex dynamic scene from its two perspective images, \emph{without object-level motion segmentation} as an essential intermediate step.




\subsection{Separate treatment for rigid SfM and non-rigid SfM}\label{ss:assumptions}
Our investigation shows that the algorithms for deformable object 3D reconstruction often differs from a rigidly moving object. Not only solutions, but even the assumptions varies significantly e.g orthographic projection, low-rank shape \cite{bregler2000recovering} \cite{dai2014simple} \cite{lee2013procrustean} \cite{kumar2017spatio}. The reason for such inadequacy is perfectly valid due to the under-constraint nature of the problem itself. This motivated us to develop an algorithm that can provide \emph{i.e} ``{\emph{ 3D reconstruction of entire dynamic scene and the non-rigidly deforming object under similar assumptions and formulation.}}''

Although to accomplish this goal for any arbitrary non-rigid deformation remains an open problem, experiments suggest that our framework under the aforementioned assumptions about the scene and the deformation, can reconstruct a general dynamic scene irrespective of the scene rigidity type. Thanks to the recent advancement in the dense optical flow algorithms \cite{bailer2015flow} \cite{chen2016full} which can reliably capture smooth non-rigid deformation over frames. These robust dense optical flow algorithms allow us to exploit local motion of deforming surfaces. Thus, our formulation is competent enough to bridge this gap between rigid and non-rigid SfM.


The main contributions of our work are as follows:
\begin{enumerate}
    \item  A framework which disentangles object-level motion segmentation for dense 3D reconstruction of a complex dynamic scene.

    \item  A common framework for dense two-frame 3D reconstruction of a complex dynamic scene (including deformable objects), which achieves state-of-the-art performance.

    \item  A new idea to resolve the inherent relative scale ambiguity problem in monocular 3D reconstruction by exploiting the as-rigid-as-possible ({\small ARAP}) constraint \cite{sorkine2007rigid}.
\end{enumerate}

\section{Outline of the Algorithm}
Before providing the details of our algorithm, we would like to introduce some common notations that are used throughout the paper.
\subsection{Notation}
We represent two consecutive images as $\mathbf{I}$, $\mathbf{I}'$ : $\Omega $ $\rightarrow \mathbb{R}^3$ $ |\Omega \subset \mathbb{Z}^2$, also referred as reference image and next image respectively. Vectors are represented by bold lower case letter, such as `${\bf{x}}$' and matrices are represented by bold upper case letter such as `${\bf{X}}$'. The subscript `a', `b' denotes anchor point and boundary point respectively, for e.g ${\bf{x_{ai}}}$, ${\bf{x_{bi}}}$ represents anchor point and boundary point corresponding to $\mathbf{i}^{th}$ superpixel in the image space. The 1-norm,  2-norm of a vector is denoted as $|.|_1$ and $\|.\|_2$ respectively. For matrices, Frobenius norm is denoted as $\|.\|_{\m F}$.

\subsection{Overview}\label{ssec:overview}
We first over-segment the reference image into superpixels, then model the deformation of the scene by a union of piece-wise rigid motions of these superpixels. Specifically, we divide the overall non-rigid reconstruction into a local rigid reconstruction of each superpixel, followed by an assembly process which glues all these individual local reconstructions in a globally coherent manner.  While the concept of the above divide-and-conquer procedure looks simple, there is however a fundamental difficulty (of {\em scale indeterminacy}) in its implementation. {Scale-Indeterminacy} refers to the well-known fact that using a moving camera one can only recover the 3D structure up to an unknown scale.  In our method, the individual rigid reconstruction of each superpixel can only be determined up to an unknown scale, the assembly of the entire non-rigid scene is only possible if and only if these scales among the superpixels are solved ---which is, however, a challenging open task itself.

In this paper, we show how this can be done using two very mild assumption \S \ref{ss:assumptions}. Under these assumptions, our method solves the unknown relative scales and obtains a globally-coherent dense 3D reconstruction of a complex dynamic scene from its two perspective views.

\subsection{Problem Statement}\label{ssec:problemstatement}
To implement the above idea of piecewise rigid reconstruction, we first partition the reference image $\mathbf{I}$ into set of superpixels $\xi_{I} = $ $\{\mathbf{s}_1, \mathbf{s}_2,.., \mathbf{s}_{\m i},.., \mathbf{s}_{\m N}\}$, where each superpixel $\mathbf{s}_{\m i} $ is parametrized by its boundary pixels $\{\mathbf{x_{bi}}=[u_\mathbf{bi}, v_\mathbf{bi},1]^{\m T} ~| \mathbf{b} = 1,..., \m B_{\m i}\}$ and an {\em anchor point} $\mathbf{x_{ai}}$ corresponding to the centroid of the $\mathbf{i}^{th}$ superpixel in the image plane. Such a superpixel partition of the image plane naturally induces a piecewise-smooth over segmentation of the corresponding 3D scene surface. We denote this set of 3D scene surfaces as $\xi_{W}$ = $\{\mathbf{\tilde{s}_{1}}, \mathbf{\tilde{s}_{2}}, ... \mathbf{\tilde{s}_{i}}, ... \mathbf{\tilde{s}}_{\m N}\}$. Although {\it{surfel}} is perhaps a better term, we nevertheless call it ``3D superpixel" for the sake of easy exposition. We further assume each 3D superpixel (`$\mathbf{\tilde{s}_{\m i}}$') is a small 3D {\em planar patch} $\Pi_\mathbf{{\tilde{si}}} =  \Big\{{\mathbf{n_{i}}}, {\mathbf{\tilde{x}_{ai}}}, \{ {\mathbf{\tilde{x}_{bi}\}}} : (\mathbf{n_{i}}, {\mathbf{\tilde{x}_{ai}}}) \in\mathbb{R}^3$ and $\{ {\mathbf{\tilde{x}_{bi}\}}} \in \mathbb{R}^{3\times \m B_{\m i}} \Big\}$, which is parameterized by surface normal $\mathbf{n_{i}}$, 3D anchor-point ${\mathbf{\tilde{x}_{ai}}}$, and 3D  boundary-points $\{ {\mathbf{\tilde{x}_{bi}\}}}$  (i.e these are the pre-images of $\mathbf{x_{ai}}$ and $\{\mathbf{x_{b i} }\}$). Assume every 3D superpixel $\mathbf{\tilde{s}_{\m i}}$ moves rigidly according $\mathbf{M}_{\m i}=
\left(
 \begin{smallmatrix}
   \mathbf{R}_{\m i} & \lambda_{\m i} {\mathbf{\hat{t}}}_{\m i}\\
   \mathbf{0} & 1
 \end {smallmatrix}
\right) \in\mathrm{SE}(3),$
where $\mathbf{R}_{\m i}$ represents relative rotation, ${\mathbf{\hat{t}}}_{\m i}$ is the translation direction, and $\lambda_{\m i}$ the unknown scale.

After our notation and symbol introduction, we are in a position to put our idea in a more precise way:  
Given two intrinsically calibrated perspective images $\mathbf{I}$ and $\mathbf{I}'$ of a generally dynamic scene and the corresponding dense optical flow field, our task is to reconstruct a piecewise-planar 
approximation of the dynamic scene surface. The deformable scene surface in the reference frame 
(i.e, $\xi_W$) and the one in the second frame (i.e, $\xi_W'$) are parametrized by 
their respective 3D superpixels $\{\mathbf{\tilde{s}_{i}}\}$ and $\{\mathbf{\tilde{s}'_{i}}\}$, where 
each ${\mathbf{\tilde{s}_{i}}}$ is described by its surface normal $\mathbf{n_{i}}$ and 
an anchor point $\mathbf{\tilde{x}_{ai}}$. Any 3D plane can be determined by an 
anchor point $\mathbf{\tilde{x}_{ai}}$ and a surface normal $\mathbf{n_i}$. If one can estimate 
correct placement of all the 3D anchor points and all the surface normals corresponding to the reference 
frame, the problem is solved, since each element of $\xi_W$ is related to $\xi'_W$ via $\mathrm{SE}(3)$ 
transformation (locally rigid).

The overall procedure of our method is presented in {\bf{Algorithm 1}}.
\begin{algorithm}[h!]
\label{Algorithm}
\caption{:~~SuperPixel Soup}
\begin{algorithmic}
\STATE {\bf{Input:}} Two consecutive image frames of a dynamic scene and dense optical flow correspondences between them.
\STATE {\bf{Output:}} 3D reconstruction for both images.
\STATE 1. Divide the reference image into '$\m N$' superpixels and construct a K-NN graph to represent the entire scene as a graph $\m G( \m V, \m E)$ defined over these superpixels \S \ref{ss:formulation}.
\STATE 2. Employ two-view epipolar geometry to recover the rigid motion and shape for each 3D superpixel \S \ref{ss:implementation}.
\STATE 3. Optimize the proposed energy function to assemble (or glue) and align all the reconstructed superpixels (``3D Superpixel Jigsaw Puzzle'') \S \ref{ssec:optimization}.
\STATE Note: The procedure of the above algorithm looks simple; there is, however, a fundamental difficulty of scale indeterminacy in its execution.
\end{algorithmic}
\end{algorithm}

\subsection{Formulation}\label{ss:formulation}
\begin{figure}
\begin{center}
\includegraphics[width=1.0\linewidth]{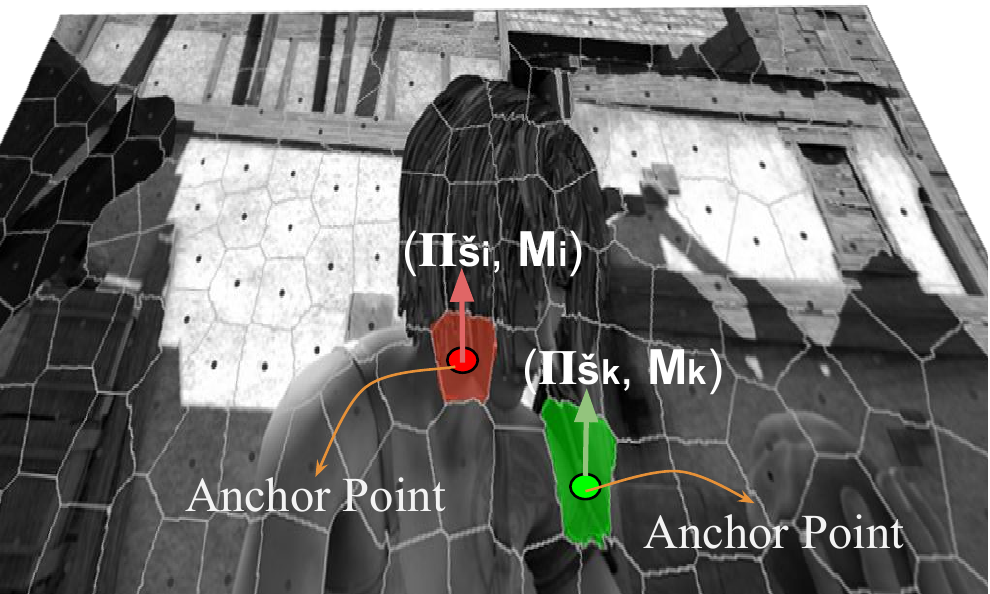}
\caption{\scriptsize Illustration shows the modeling of a continuous scene with a piece wise rigid and planar assumption. Each superpixel is composed of a set $(\Pi_{\mathbf{\tilde{si}}}, \mathbf{M_i})$ where $\Pi_{\mathbf{\tilde{si}}}$ contains geometric parameters such as normal, anchor point, boundary points of a plane in 3D and $\mathbf{M_i}$ contains the motion parameters i.e rotation and translation. \label{fig:symbolIntro1}}
\end{center}
\end{figure}

We begin by briefly reiterating some of our representation. 
We partition the reference image into a set $\xi_I$, whose corresponding set in the 
3D world is $\xi_W$. Equivalently, $\xi_I'$ and $\xi_W'$ are the respective sets for the next 
frame. The mapping of each element in the reference frame and next frame differs by a 
rigid transformation. Mathematically, $\xi_W \mapsto \xi_W'$ via $\mathrm{SE}(3)$ transformation 
(also known as special euclidean group), for instance $\mathbf{\tilde{x}_{ai}'}$ = $\mathbf{M_i}\mathbf{\tilde{x}_{ai}}$ 
where $\mathbf{\tilde{x}_{ai}'} \subset \xi_{W}'$ and $\mathbf{\tilde{x}_{ai}} \subset \xi_{W}$.  In our formulation each 
3D plane is described by $\Phi_\mathbf{{\tilde{si}}}$ = \Big\{($\Pi_\mathbf{{\tilde{si}}}$, $\mathbf{M_i}$) $|$ $\forall$ $\mathbf{i} \in [1, \m N]$\Big\}, where $\m N$ is the total number of 
superpixels (see Fig. \ref{fig:symbolIntro1}). Similarly, in the image space $\xi_I \mapsto \xi_I'$ through the plane-induced homography $\mathbf{s_i}'= \mathbf{K} \left(\mathbf{R}_{\m i}-\frac{\lambda_{\m i} \mathbf{{t}}_{\m i} \mathbf{n}_{\m i}^{\m T}}{\lambda_{\m i} \m d_{\m i}}\right)\mathbf{K}^{-1}\mathbf{s_i}$ \cite{hartley2003multiple}\footnote{scale $\lambda_{\m i}$ is introduced 
both in the numerator and denominator for clarification that scale does not affect the homography transformation.}. Here, $\mathbf{K}$ is the 
intrinsic camera matrix and $\m d_{\m i}$ is the depth of the plane. Using these notations and definitions, we build a K-NN graph.

\begin{figure}[t]
\begin{center}
\includegraphics[width=1.0\linewidth]{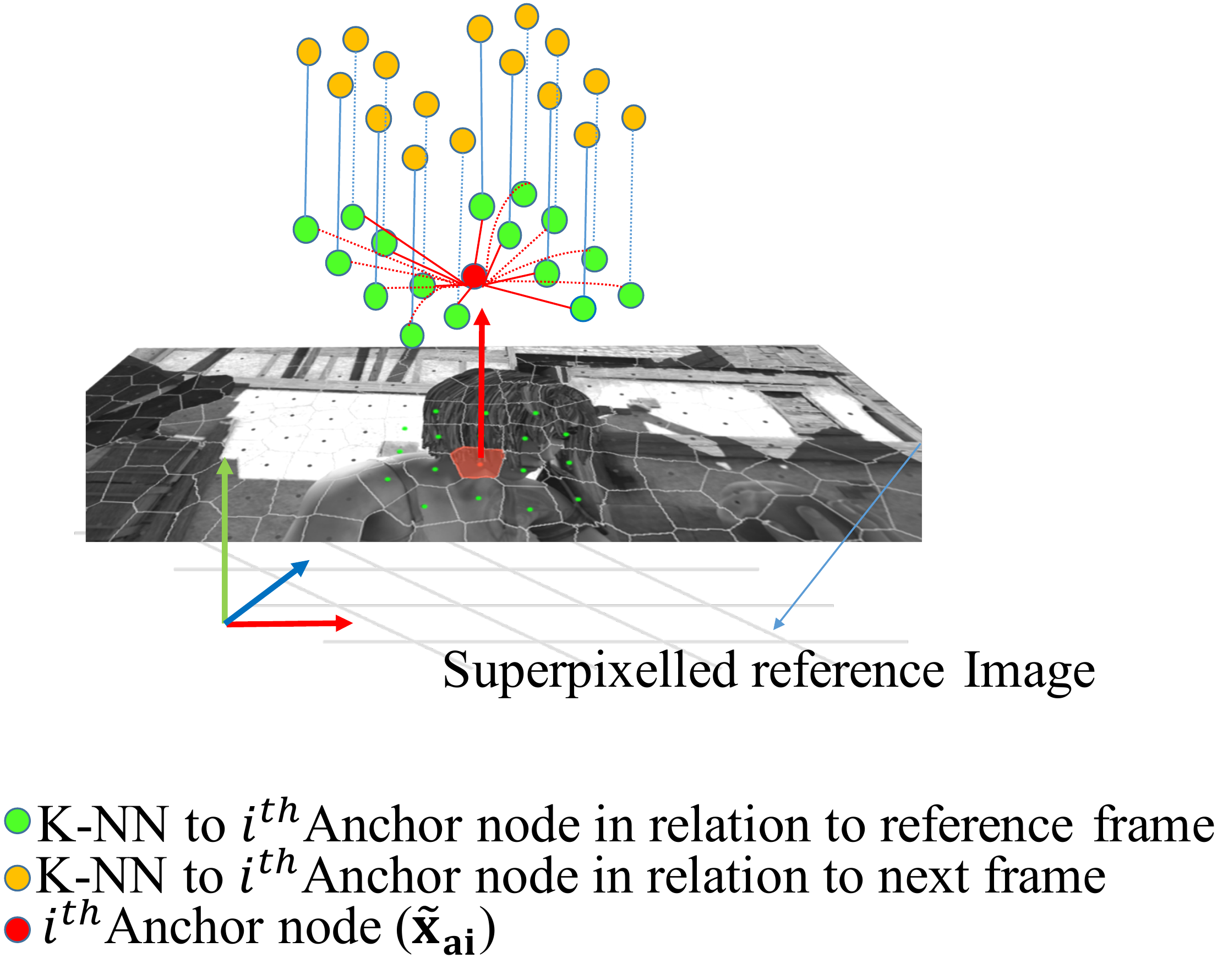}
\caption{\scriptsize Demonstration of as rigid as possible constraint. Superpixel segmentation in the reference frame is used to decompose the entire scene as a set of anchor points. Schematic representation shows the construction of K-NN around a particular anchor point (shown in Red). We constrain the local 3D coordinate transformation both before and after motion (green shows K-NN the reference frame, yellow shows the relation in the next frame (after motion))\label{fig:ARAPKNN}. We want this transformation to be as rigid as possible.}
\end{center}
\end{figure}

{\bf{Build a K-NN graph}}: Using over-segmentation of the reference image $\xi_I$ (which is the projection of a set of 3D planes $\Phi_\mathbf{{\tilde{si}}}$) and Euclidean distance metric, we 
construct a K-NN graph $\m G(\m V, \m E)$ in the \emph{image space} connecting each anchor point to its K-nearest anchor points.
The graph vertices ($\m V$) are composed of anchor point that connects to other anchor points via graph edges ($\m E$). The distance between any two vertices ($\m E_{\m i}\subset\m E$) is taken as the Euclidean distance between them. Here, we assume Euclidean distance as a 
valid graph metric to describe the edge length between any two local vertices. 
Such an assumption is valid for local compactness (Euclidean spaces are locally compact). Interested reader may refer to \cite{burago2001course} \cite{williamson1987constructing} \cite{whiteley2004rigidity} for 
comprehensive details. Here, 'K' is the number of nearest neighbor that is used to construct local graph structure. This K-NN graph relation helps to constrain the motion and continuity of the space (defined in terms of optical flow, depth). To impose a hard constraint, we build a K-NN graph using anchor point beyond its immediate neighbors (Fig. \ref{fig:ARAPKNN}).

This K-NN graph is crucial in the establishment of local rigidity constraint which is the basis of our assumption. This graph structure allows us to enforce our assumption \emph{i.e}, the shape to be as rigid as possible globally and rigid locally.

{\bf{As-Rigid-As-Possible (ARAP) Energy Term}}: Our method is built on the idea that 
the correct scales of 3D superpixels can be estimated by enforcing prior assumptions that govern the deformation of the dynamic surface.  Specifically, we require that, locally, the motion that each 3D-superpixel undergoes is rigid, and globally the entire dynamic scene surface must move as rigid as possible (ARAP). In other words, while the dynamic scene is globally non-rigid, its deformation must be {\em regular} in the sense that it deforms as rigidly as possible.  To implement this idea, we define an ARAP-energy term as:
\begin{equation}\label{eq:EARAP}
{\begin{aligned}
& \displaystyle \m E_{\textrm{arap}}= \\
& \displaystyle \sum_{\mathbf{i}=1}^{\m N} \sum_{\mathbf{k} \in \mathcal{N}_\mathbf{i}} w_{1}(\mathbf{x_{ai}}, \mathbf{x_{ak}}) \Big(\|\mathbf{R_{i}}- \mathbf{R_{k}}\|_{\m F} + \|\lambda_{\m i}\mathbf{\hat{t}_i}- \lambda_{\m k}\mathbf{\hat{t}_{k}}\|_{\m 2}\Big) \\
& + \displaystyle w_{2}(\mathbf{x_{ai}}, \mathbf{x_{ak}}).\Big| \|{\mathbf{\tilde{x}_{ai}}}- {\mathbf{\tilde{x}_{ak}}}\|_2 - \|{\mathbf{\tilde{x}_{ai}'}}- {\mathbf{\tilde{x}_{ak}}'}\|_2\Big|_1.
\end{aligned}}
\end{equation}
Here, the first term favors smooth motion between the local neighbors, while the second term encourages inter-node distances between the anchor node and its K nearest neighbor nodes (denoted as $\mathbf{k} \in \mathcal{N}_\mathbf{i}$) to be preserved before and after motion (hence as-rigid-as-possible, see Fig. \ref{fig:ARAPKNN}). We define the weighting parameters as:
\begin{equation}\label{eq:wc} w_{1}(\mathbf{x_{ai}}, \mathbf{x_{ak}}) = w_{2}(\mathbf{x_{ai}}, \mathbf{x_{ak}}) = \exp(- \beta \|\mathbf{x_{ai}}- \mathbf{x_{ak}} \|).
\end{equation}
These weights are set to be inversely proportional
 to the distance between two superpixels. 
 This is to reflect our intuition that, 
 the further apart two superpixels are, the weaker the $\m E_{\textrm{arap}}$ energy is. Although there may be 
 redundant information in these two terms w.r.t scale estimation, we keep them 
 for motion refinement \S \ref{ssec:optimization}. Note that, this term is only defined over anchor points, hence it enforces no depth smoothness along boundaries. The weighting term in $\m E_{\textrm{arap}}$ advocates the local rigidity by penalizing over the distance between anchor points. This allows immediate neighbors to have smooth deformation over time. Also, note that $\m E_{\textrm{arap} }$ is generally {\em non-convex}. This non-convexity is due to the second term in Eq. \ref{eq:EARAP}, where we have a minus sign between two $l_2$ norm terms. In Eq. \ref{eq:wc} $\beta$ is an empirical constant.

$\m E_\textrm{arap}$ alone is good enough to provide reasonably correct scales, however, the piece-wise planar composition of a continuous 
3D space creates discontinuity near the boundaries of each plane. For this reason, we incorporate additional constraint to fix this depth discontinuity and further refine motions and geometry for each superpixel via neighboring relations. We call these constraints as Planar Re-projection, 3D Continuity and Orientation Energy constraint.

{\bf{Planar Re-projection Energy Term}}:
With the assumption that each superpixel represents a plane in 3D, it must satisfy corresponding planar reprojection error in 2D image space. This reprojection cost reflects the average dissimilarity in the optical flow correspondences across the entire superpixel due to motion. Therefore, it helps us to constrain the surface normal, rotation and translation direction such that they obey the observed planar homography in the image space. To infer any pixel inside a superpixel, we use the operator $\psi(.)$, for e.g $\psi(\mathbf{s}_1^{j})$ will give the coordinates of $j^{th}$ pixel inside $\mathbf{s}_1$. Using it we define

\begin{equation}\label{eq:planarerror}
\begin{aligned}
& \displaystyle \m E_{\mathrm{proj}} = \\
& \displaystyle \sum_{\mathbf{i}=1}^{\m N} \frac{w_{3}}{|\psi(\mathbf{s_i})|}\sum_{\mathbf{j}=1}^{|\psi(\mathbf{s_i})|}\|\psi(\mathbf{s_{i}^{j}})'- \mathbf{K}(\mathbf{R_{i}}-\frac{\mathbf{t_{i}} \mathbf{n_{i}}^{\m T}}{\m d_{\m i}})\mathbf{K}^{-1} \psi(\mathbf{s_{i}^j}) \|_2.
\end{aligned}
\end{equation}
Here, $\psi(\mathbf{s_{i}^{j}})$, $\psi(\mathbf{s_{i}^{j}})'$ is the optical flow correspondence of $\mathbf{j}^{th}$ pixel inside $\mathbf{i}^{th}$ superpixel in the reference frame and next frame respectively. The operator $|.|$ represent the cardinal number of a set. $w_3$ is a trade-off scalar chosen empirically. A natural question that may arise is: {\emph{This term is independent of scale, then what's the purpose of this constraint? How does it help?}} Kindly, refer to \S \ref{ssec:optimization} for details.

{\bf{3D Continuity Energy Term}}:
In case of a dynamic scene, where both camera and the objects are in motion, its quite apparent that the scene will undergo some changes across frames. Hence, to assume unremitting global continuity with a piece-wise planar assumption, in a dynamic scene is unreasonable. Instead, local weak continuity constraint can be enforced ---a constraint that can be broken occasionally \cite{hinton1977relaxation} i.e., local planes are connected to few of its neighbors. Accordingly, we want to allow local neighbors to be piece-wise continuous. To favor this continuous or smooth surface reconstruction, we require neighboring superpixels to have a smooth depth transition at their boundaries. To do so, we define a 3D continuity energy term as:
\begin{equation}\label{eq:depthcont}
\begin{aligned}
& \displaystyle \m E_{\mathrm{cont}}= \\
& \displaystyle \sum_{\mathbf{i}=1}^{\m N} \sum_{\mathbf{k} \in \mathcal{N}_\mathbf{i}} w_4(\mathbf{X_{bi}}, \mathbf{X_{bk}}) \left(\|\mathbf{\tilde{X}_{bi}}- \mathbf{\tilde{X}_{bk}}\|_{\m F} + \rho(\|\mathbf{\tilde{X}_{bi}}'-\mathbf{\tilde{X}_{bk}}'\|_{\m F}\right)
\end{aligned}
\end{equation}

\begin{figure}[t]
\begin{center}
\includegraphics[width=1.0\linewidth, height=0.6\linewidth]{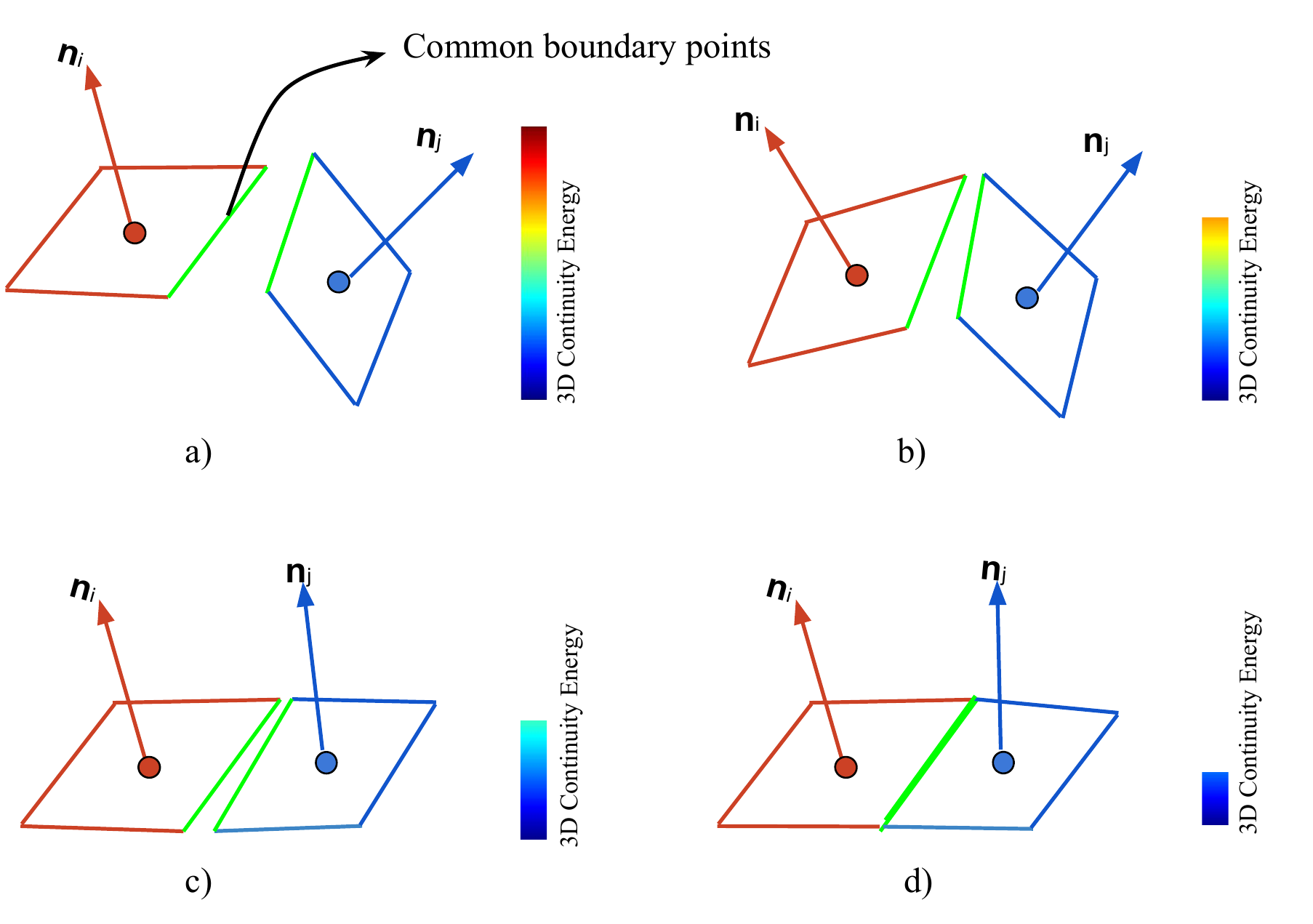}
\caption{\scriptsize 3D Continuity energy favors continuous surface for the planes that shares the common boundary points. a)-d) The lesser the $\m E_\textrm{cont}$ is, smoother the surface becomes (color bar shows the energy).  \label{fig:depthContinuity}}
\end{center}
\end{figure}

where, $\mathbf{X}$, $\mathbf{\tilde{X}}$ represents the corresponding matrices in 2D image space and 3D Euclidean space ($\mathbf{X_{bi}} \in \mathbb{R}^{2 \times{\m B \m i}}, \mathbf{\tilde{X}_{bi}} \in \mathbb{R}^{3\times{\m B \m i}}$, where $\m B \m i$ is the total number of boundary pixel for $\mathbf{i}^{th}$ superpixel).  Since in our representation, geometry and motion are shared among all pixels within a superpixel, so regularization within the superpixel is not explicitly needed. Thus, we only concentrate on the shared boundary pixels to regularize our energy. Note that the neighboring relationship in $\m E_\textrm{cont}$ is different from $\m E_{\mathrm{arap} }$ term.  Here, the neighbors share common boundaries with each other.

To encourage the geometry to be approximately smooth locally if the object has similar appearance, we color weight the energy term along the boundary pixels. For each boundary pixel of a given superpixel, we consider its 4-connected neighboring pixels to weight. Using this idea for $w_4$ we obtain:
\begin{equation}\label{eq:w4}
w_{4}(\mathbf{X_{bi}}, \mathbf{X_{bk}}) = \sum_{j=1}^{4} \exp(-\beta \|\mathbf{I}(\mathbf{X_{bi}}) - \mathbf{I}({\zeta_j}) \|_{\m F})
\end{equation} which weigh the inter-plane transition by color difference. The symbol $\zeta_j\in\mathbb{R}^{2\times \m B \m i}$ is a set that contains the 4 connecting pixels to each $\mathbf{i}^{th}$ superpixel boundary pixel shared with $\mathbf{k}^{th}$ superpixel. The color based weighting term plays an important role to allow for ``weak continuity constraint'' i.e gradually allow for occasional discontinuity \cite{hinton1977relaxation} \cite{blake1983least}.

To better understand the implication of $\m E_{\mathrm{cont}}$ constraint, consider two boundary points in the image space $a, b \in \mathbb{R}^2$. Generally, if these two points lie on a different plane, it will not coincide in the 3D space before and after motion. Hence, we compute the 3D distance between boundary pixels corresponding to both reference frame and next frame, which leads to our goal of penalizing distance along shared edges (see Fig. \ref{fig:depthContinuity}). Therefore, this term ensures the 3D coordinates across superpixel boundaries to be continuous in both frames.
The challenge here is to reach a satisfactory solution for overall scene continuity, almost everywhere in both the frames \cite{blake1987visual}.
In the Eq. \ref{eq:depthcont} $\rho$ is a truncation function defined as $\rho=\min(., \sigma)$ and similar to Eq. \ref{eq:wc} $\beta$ in Eq. \ref{eq:w4} is a constant, chosen empirically.

{\bf{Orientation Energy Term}}: To encourage the smoothness in the orientation of the neighboring planes, we added one more geometric constraint {\emph{i.e,}} $\m E_\mathrm{orient}$ defined as follows.
\begin{equation}
\m E_\textrm{orient} =  \sum_{\mathbf{i}=1}^{\m N} \sum_{\mathbf{k} \in \mathcal{N}_\mathbf{i}} \rho_n \Big(1-\mathbf{n_{i}}^{\m T} \mathbf{n_{k}}\Big)
\end{equation}\label{eq:Eorient}
Here neighbor index is same as 3D continuity term. $\rho_n$ denotes the truncated $l_1$ penalty function which is defined as $\rho_n(x) = \min(|x|, n)$. Intuitively, it encourages the similarity between neighboring normal's and truncate any value more than $n$.

{\bf{Combined Energy Function}}:  Equipped with all these constraints, we define our overall cost function or energy function to obtain a scale consistent 3D reconstruction of a complex dynamic scene. Our goal is to estimate depth ($\mathbf{d_i}$), surface normal ($\mathbf{n_i}$) and scale $\lambda_{\mathbf{i}}$ for each 3D planar superpixel. The key is to estimate the unknown relative scale $\lambda_\mathbf{i}$.  We solve this by minimizing the following energy function:
\begin{equation}
\begin{aligned}
& \displaystyle \underset{{\lambda_\mathbf{i}}, {\mathbf{n_i}}, \mathbf{d_i}, \mathbf{R_i}, \mathbf{t_i}} \min  \hspace{.1cm} \m E = \m E_{\mathrm{arap}} +\alpha_1 \m E_{\mathrm{proj}} + \alpha_2 \m E_{\mathrm{cont} } + \alpha_3 \m E_\mathrm{orient}  \\
& \displaystyle \text{{subject to}} \displaystyle{\sum_{i=1}^{\m N} \lambda_\mathbf{i}= 1, \lambda_\mathbf{i}>0}.\\
& \displaystyle \mathbf{R_i} \in \mathbb{SO}(3), \|\mathbf{n_i}\|_2=1.
\end{aligned}\label{eq:CostFunction}
\end{equation}

The equality constraint on $\lambda$ fixes the unknown freedom of a global scale. The constraint on $\mathbf{R_i}$ is imposed to restrict the rotation matrix to lie on $\mathbb{SO}(3)$ manifold. In our formulation, the rotation matrix represents the combined Euler 3D angles. Although there are other efficient representations for 3D rotation, we used  $\mathbf{R_i} \in \mathbb{R}^{3\times 3}$ matrix representation as it comes naturally via epipolar geometric constraint, hence, further post-conversion steps can be avoided. The constant $\alpha_1, \alpha_2, \alpha_3$ are included for numerical consistency.

\subsection{Implementation}\label{ss:implementation}
We partition the reference image into 1,000-2,000 superpixels \cite{achanta2012slic}. Parameters such as $\alpha_1$, $\alpha_2$, $\alpha_3$, $\beta$, $\sigma$ were tuned differently for different datasets. To perform optimization of the proposed energy function (Eq. \ref{eq:CostFunction}), we require initial set of proposals for motion and geometry.

\begin{figure}
\centering
\includegraphics[width=0.50\textwidth] {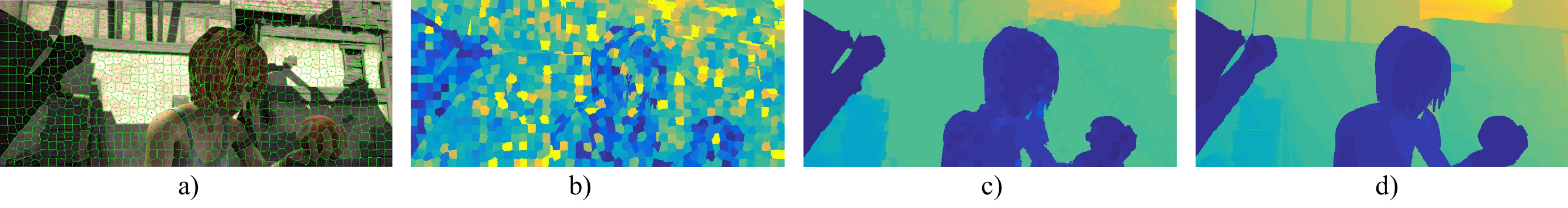}
\caption{ \scriptsize a) Superpixelled reference image b) Individual superpixel depth with arbitrary scale ({\it{unorganised superpixel soup}}) c) recovered depth map using our approach ({\it{organised superpixel soup}}) d) ground-truth depth map.}
\label{fig:solutionExample}
\end{figure}

\subsubsection{Initial Proposal Generation} \label{sec:varinit}
We exploit piece-wise rigid and planar assumption to estimate an initial proposal for geometry and motion. We start by estimating homography $(\mathbf{H}_{\pi \m i})$ for each superpixel using dense feature correspondences. Piece-wise rigid assumption helps in approximate estimation of rotation and correct translation direction via triangulation and chierality check \cite{hartley2003multiple} \cite{hartley1997triangulation}. To obtain the correct normal direction and initial depth estimate, we solve the following set of equations for each superpixel:
\begin{equation}
\begin{aligned}
& \displaystyle \mathbf{H}_{\pi \m i} =  \mathbf{K}(\mathbf{R_{i}}-\frac{\mathbf{t_{i}} \mathbf{n_{i}}^{\m T}}{\m d_{\m i}})\mathbf{K}^{-1}
\end{aligned}
\end{equation}
The reason we choose this strategy to obtain normal is because a simple decomposition of homography matrix to the rotation, translation and normal can lead to sign ambiguity \cite{varol2009template} \cite{malis2007deeper}. Nevertheless, if one has correct rotation and direction of translation --which we infer from  chierality check, then inferring normal becomes easy\footnote{ \scriptsize The solution to the obtained normal must be normalized.}. Here, we assume the depth '$\m d_{\m i}$' to be a positive constant and the initial arbitrary reconstruction is in the +Z direction. This strategy of gathering 9-dimensional variables (6-motion variable and 3-geometry variable) for each individual superpixel gives us a good enough estimate to get started with the minimization of our overall energy function \footnote{\scriptsize If the size of the superpixel is very small, use the neighboring superpixels optical flow to estimate motion parameters.}.

To initialize 3D vectors in our formulation we use the following well known relation:
\begin{equation}
\mathbf{\tilde{x}_{ai}}=\Bigg[\big(\frac{u_\textrm{ai} - c_x}{f_x}\big), \big(\frac{v_\textrm{ai} - c_y}{f_y}\big), 1 / \mathbf{n_{i}}^{\m T} \mathbf{K}^{-1} \begin{pmatrix}
 u_\textrm{ai} \\
 v_\textrm{ai} \\
 1
\end{pmatrix}
\Bigg]^{\m T} (\lambda_{\m i} \m d_{\m i})
\end{equation}
where, $(u_\textrm{ai}, v_\textrm{ai})$ are image coordinates and $(c_x, c_y, f_x, f_y)$ are camera intrinsic parameters which can be inferred from $\mathbf{K}$ matrix.

\subsubsection{Optimization}\label{ssec:optimization}
With good enough initialization of the variables, we start to optimize our energy function Eq. \ref{eq:CostFunction}.  A global optimal solution is hard to achieve due to the non-convex nature of the proposed cost function (Eq. \ref{eq:CostFunction}).
However, it can be solved efficiently using interior-point methods \cite{benson2002interior} \cite{benson2014interior}.  Although the 
solution found by the interior point method is at best local minimizer, empirically they appear to give good 3D reconstruction.  In our experiments, we initialized all $\lambda$'s with an initial value of $\frac{1.0}{\m N}$.

Next, we employ a particle based refinement algorithm to rectify our initial motion and geometry beliefs. Specifically, we used the Max-Product Particle Belief propagation (MP-PBP) procedure with the TRW-S algorithm \cite{kolmogorov2006convergent} to optimize over the surface normals, rotations, translations and depths for all 3D superpixels using Eq. \ref{eq:Eref}.  We generated 50 particles as proposals for the unknown parameters around the already known beliefs to initiate refinement moves.  Repeating this strategy for 5-10 iterations, we obtain a smooth and refined 3D structure of the dynamic scene.
 \begin{equation}
 \m E_\textrm{ref} = \m E_{\mathrm{arap}} + \alpha_1 \m E_{\mathrm{proj}} + \alpha_2 \m E_{\mathrm{cont}} + \alpha_3 \m E_{\mathrm{orient}}. \label{eq:Eref}
 \end{equation}

\section{Experiments and Results}
We evaluated our formulation both qualitatively and quantitatively on various standard benchmark datasets, namely MPI Sintel \cite{butler2012naturalistic}, KITTI \cite{geiger2013vision}, VKITTI \cite{gaidon2016virtual} and You-Tube Object dataset \cite{prest2012learning}. All these 
dataset contains images of dynamic scene  where both camera and objects are in motion w.r.t each other. To test the reconstruction result on deformable 
objects we used Paper, T-shirt \cite{varol2009template} \cite{varol2012constrained} and Back Sequence \cite{garg2013dense}.
For evaluating the result, we selected the most commonly used error metric \emph{i.e,} mean relative error metric.

\textbf{Evaluation Metric}:
To keep the evaluation metric consistent with the previous work \cite{ranftl2016dense}, we used mean relative error (MRE) metric for evaluation. MRE is defined as $\frac{1}{\m P}\sum_{\m i=1}^{\m P}| \m z_\textrm{gt}^{\m i} - \m z_\textrm{est}^{\m i}|/ \m z_\textrm{gt}^{i}$. Here, $z_\textrm{est}^{\m i}$, $z_\textrm{gt}^{\m i}$ denotes 
the estimated and ground-truth depth respectively with $\m P$ being the total number of points.  The error is 
computed after re-scaling the recovered depth properly as the reconstruction is obtained up to an unknown global scale. Quantitative evaluation 
for the YouTube-Objects dataset and the Back dataset are missing due to the absence of ground-truth result.

To show that our same formulation works well for both rigid and non-rigid cases, 
we evaluated our method with different types of scene that contain \emph{rigid, non-rigid, complex dynamic scene i.e., composition of both rigid and non-rigid}.

\subsection{Experimental Setup and Results}

\textbf{Experimental setup and processing time}:
We partition the reference image using SLIC superpixels \cite{achanta2012slic}. We used the current state-of-the-art optical flow algorithm to compute dense optical flow \cite{bailer2015flow}. 
To initialize the motion and geometry variables, we used the the procedure discussed in \S \ref{sec:varinit}. Interior 
point algorithm \cite{benson2002interior} \cite{benson2014interior} and TRW-S \cite{kolmogorov2006convergent}  were employed to solve the proposed optimization. We 
implemented our algorithm in MATLAB/C++. Our modified implementation (modified from our ICCV implementation\cite{kumar2017monocular}) takes 
an average of 15-20 minutes to provide the result for images of size $1024 \times 436$. The processing time is estimated on a regular desktop with Intel core i7 processor (16 GB RAM) for 50 refinement particle per superpixel.

\textbf{Results on MPI Sintel Dataset}:
We begin our analysis of experimental results with MPI Sintel dataset \cite{butler2012naturalistic}. This dataset is derived from an animation movie featuring complex scenes. It contains highly dynamic sequences with large motions, significant illumination changes, and non-rigidly moving objects. This dataset has emerged as a standard benchmark to evaluate dense optical flow algorithm's and recently, it has also been used in the evaluation of dense 3D reconstruction methods for a general dynamic scene \cite{ranftl2016dense}.

\begin{figure}
\centering
\includegraphics[width=0.49\textwidth] {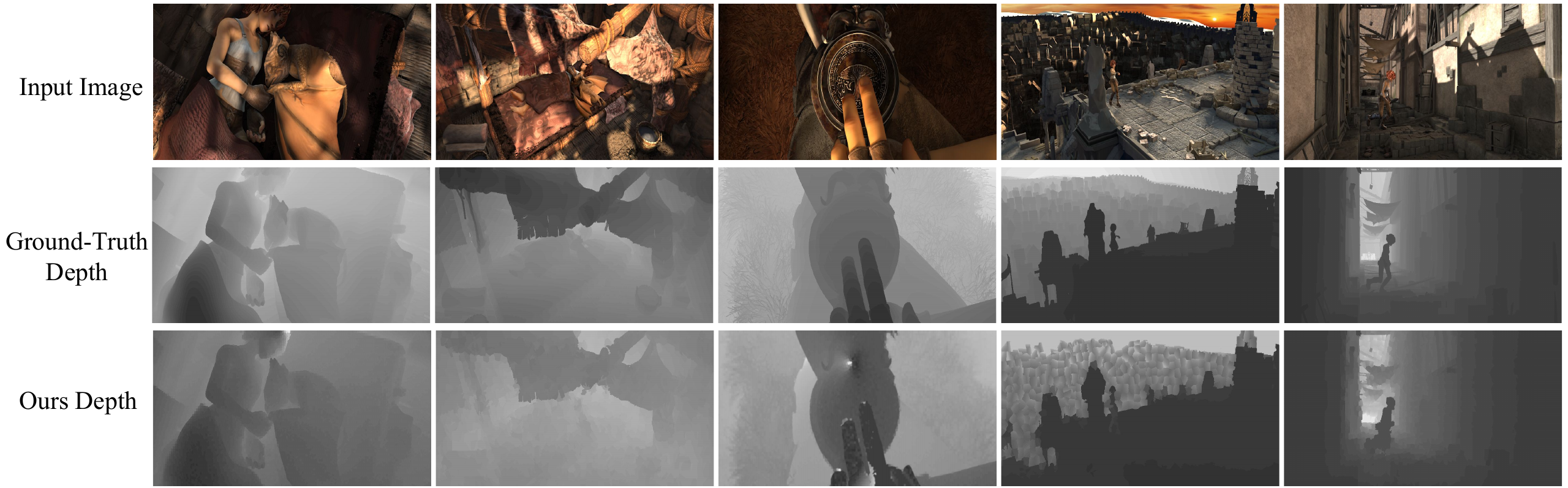}~~~
\caption{\scriptsize Qualitative results using our algorithm in a complex dynamic scene. Example images are taken from MPI Sintel dataset \cite{butler2012naturalistic}. {\bf{Top row:}}  Input reference image from {\it{sleeping\_1, sleeping\_2, shaman\_3, temple\_2, alley\_2}} sequence (from left to right). {\bf{Middle row:}} Ground-truth depth map for the respective frames. {\bf{Bottom row:}} Recovered depth map using our method.}
\label{fig:MPI-Results}
\end{figure}

The presence of non-rigid objects in the scene makes it a prominent choice for us to test our algorithm. It is a challenging dataset particularly for the piece-wise planar assumption due to the presence of many small and irregular shapes in the scene. Additionally, the presence of ground-truth depth map makes quantitative analysis much easier. We selected 120 pair of images to test our method that includes images from {\em alley\_1,  ambush\_4, mountain\_1, sleeping\_1 and temple\_2}.  Fig. \ref{fig:MPI-Results} shows some qualitative results on a few images taken from the sub-group of MPI Sintel dataset.

\begin{figure}
\centering
\includegraphics[width=0.49\textwidth] {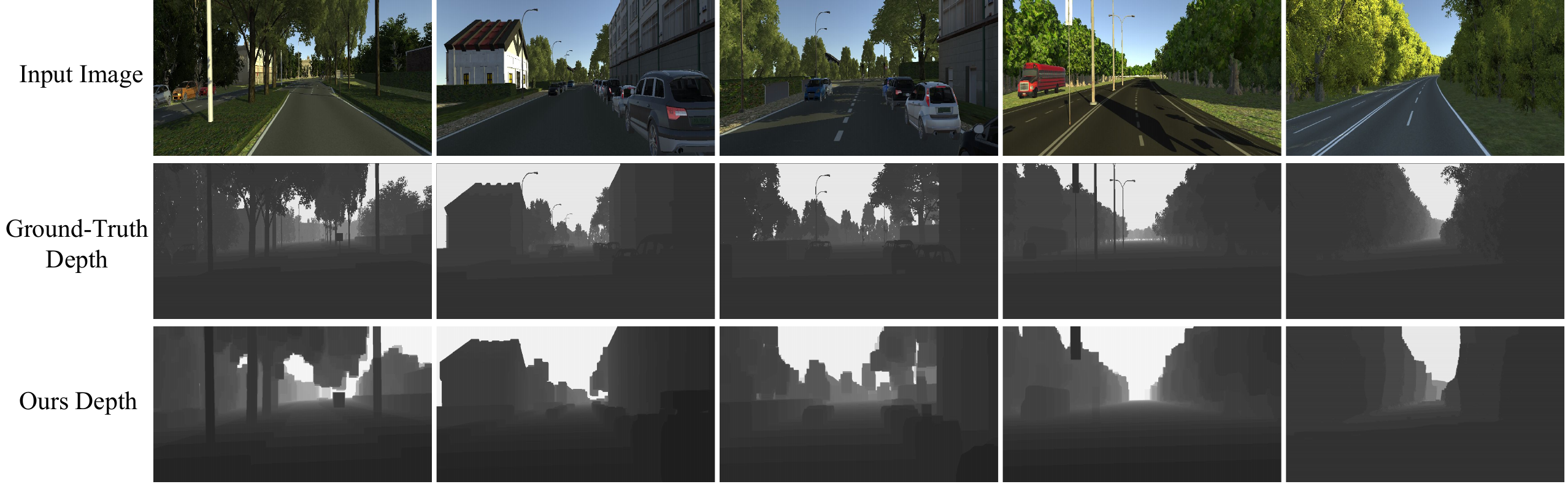}~~~
\caption{\scriptsize Qualitative results using our algorithm for the outdoor scenes. Examples are taken from VKITTI dataset \cite{gaidon2016virtual}. {\bf{Top row:}}  Input reference image. {\bf{Middle row:}}  Ground-truth depth map for the respective frames. {\bf{Bottom row:}} Recovered depth map using our method.}
\label{fig:VKITTI-Results}
\end{figure}

\textbf{Results on VKITTI Dataset}:
The Virtual KITTI dataset \cite{gaidon2016virtual} contains computer-rendered photo-realistic outdoor driving scenes which resemble KITTI dataset. The advantage of using this dataset is that it provides perfect ground-truths for many measurements.  Furthermore, it helps to simulate algorithm related to dense 3D reconstruction with distortion-free and noise-free images, 
facilitating quick experimentation.  We selected 120 pair of images from {\em 0001\_morning, 0002\_morning, 0006\_morning and 0018\_morning}. Our qualitative results in comparison to the ground-truth depth map are shown in Fig. \ref{fig:VKITTI-Results}.

\begin{figure}
  \centering
  \includegraphics[width=0.49\textwidth] {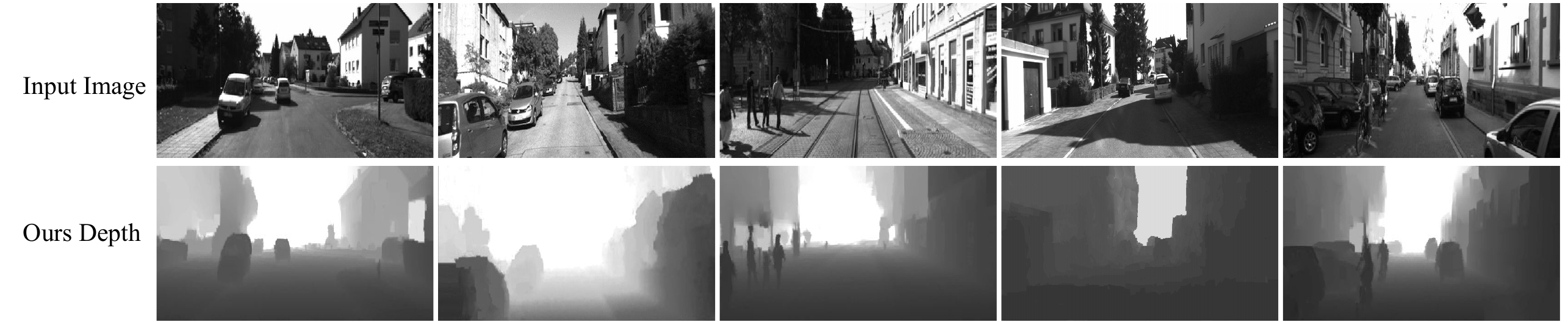}
  \caption{\scriptsize Qualitative results on KITTI Dataset \cite{geiger2013vision}. The second row shows the obtained depth map for the respective frames. Note: Dense ground-truth depth data is not available with this dataset.}
  \label{fig:KITTI-Results}
  \end{figure}

\textbf{Results on KITTI Dataset}:
The KITTI dataset \cite{geiger2013vision} features the real-world outdoor scene targeting autonomous driving application. 
The KITTI images are taken from the camera mounted on top of a car. It's a 
challenging dataset as it contains scenes with large camera motion and realistic lighting condition. In contrast to the aforementioned datasets, it only contains sparse ground-truth 3D information which makes evaluation a bit strenuous. Nonetheless, it captures noisy real-world situation and therefore, it is well suited to test the 3D reconstruction algorithm for a general dynamic scene case. We selected 00-09 sub-category from odometry dataset to evaluate and compare our results. We calculated mean relative error only over the provided sparse 3D LiDAR points --after adjusting the global scale.  Fig. \ref{fig:KITTI-Results} shows some qualitative results on few images.

\begin{figure}
  \centering
  \includegraphics[width=0.49\textwidth] {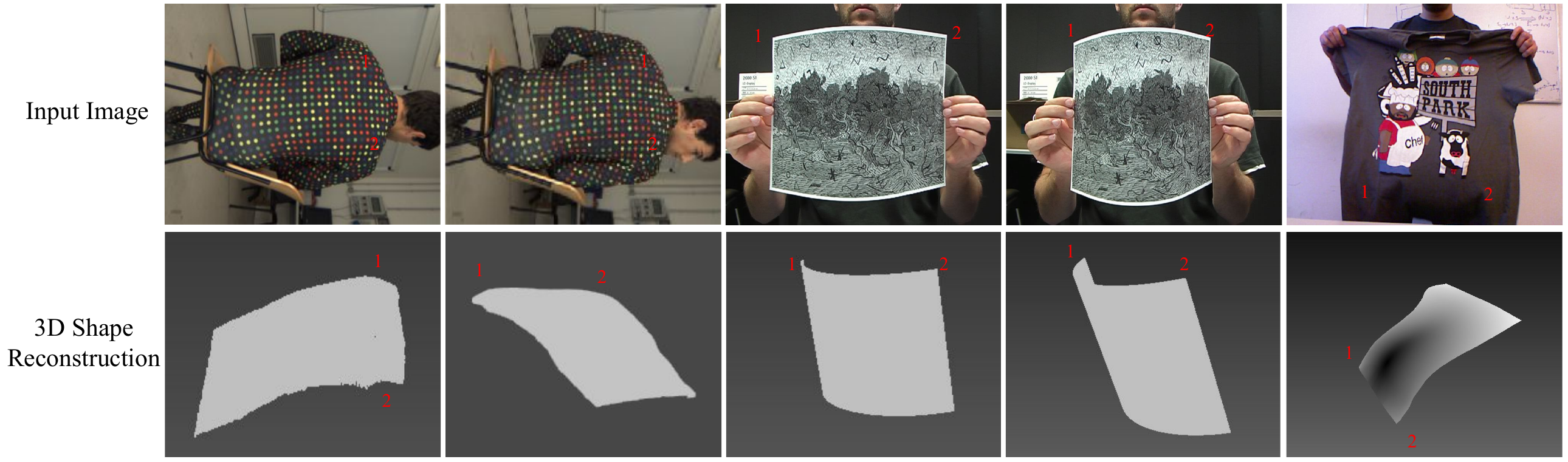}
  \caption{ \scriptsize Dense 3D reconstruction of  the objects that are undergoing non-rigid deformation over frames. {\bf{Top row:}} Input reference frame from {\it{Back sequence}} \cite{garg2013dense}, {\it{Paper sequence}} \cite{varol2009template}\cite{varol2012constrained} and {\it{t-shirt sequence}}\cite{varol2009template}\cite{varol2012constrained}.  {\bf{Bottom row:}} Qualitative 3D reconstruction results for the respective deforming object.}
  \label{fig:NonRigid-Results}
\end{figure}

\textbf{Results on Non-Rigid Sequence}
We also tested our method on some commonly 
used dense non-rigid sequence namely \emph{kinect\_paper} \cite{varol2009template}, \emph{kinect\_tshirt} \cite{varol2009template} and \emph{back sequence} \cite{garg2013dense}\footnote{Note: The intrinsic matrix for back sequence is not available with the dataset, we estimated an approximate value of it using 2D-3D relation available from Garg {\it{et. al.}}  \cite{garg2013dense}.}. Most of the benchmark approach to solve non-rigid structure from motion use multiple frames and orthographic camera model. Despite a two-frame method and perspective camera model, we are able to capture the deformation of non-rigid object and achieve its reliable reconstruction. Qualitative results  for dense non-rigid object sequence are shown in Fig. \ref{fig:NonRigid-Results}. To compute the mean relative error, we align and scale our shape (fixing global ambiguity) w.r.t ground-truth shape.

\subsection{Comparison}
\begin{figure}
\centering
\includegraphics[width=0.49\textwidth] {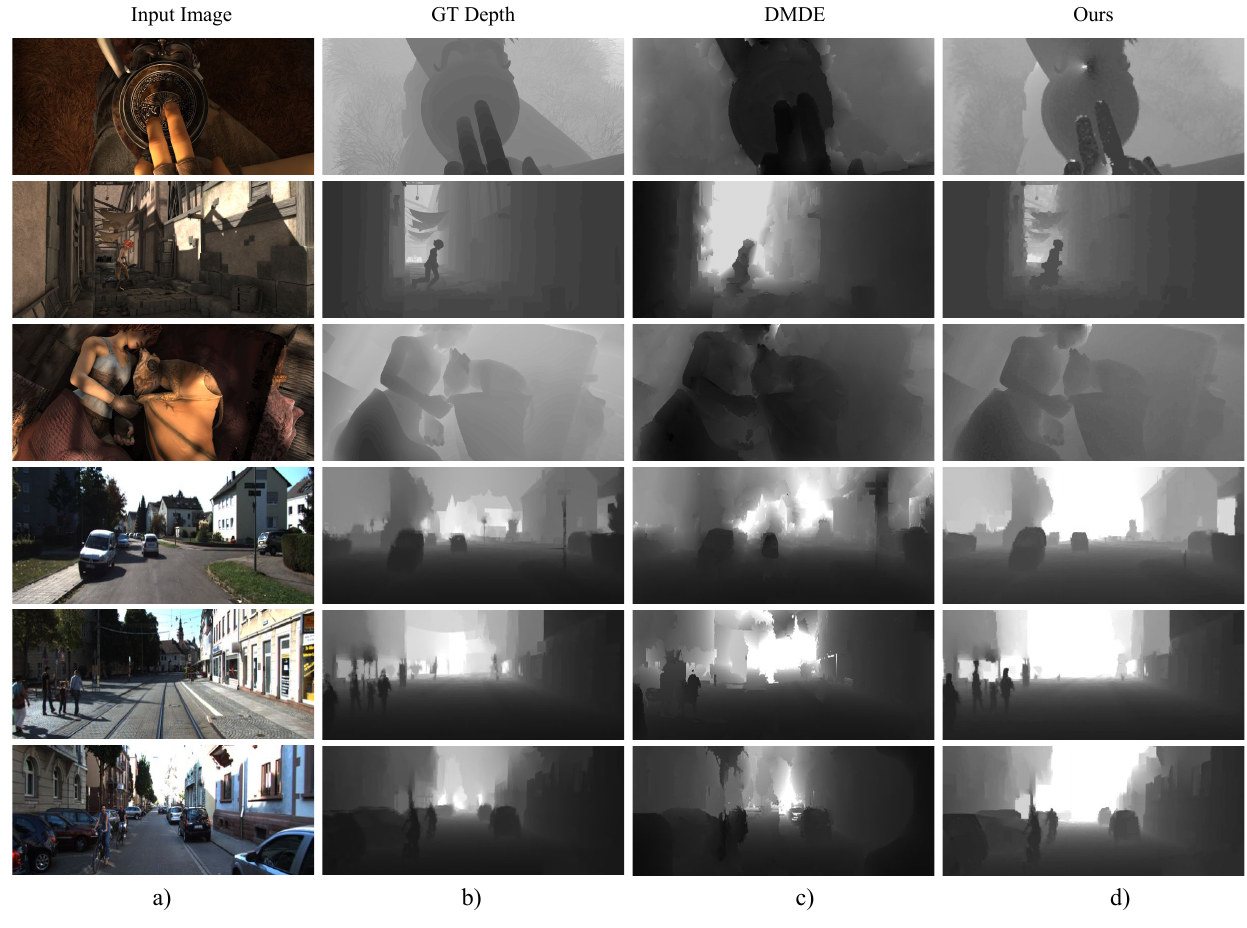}~~~
\caption{\scriptsize Qualitative comparison of our method with DMDE \cite{ranftl2016dense} on MPI Sintel \cite{butler2012naturalistic} and KITTI Dataset \cite{butler2012naturalistic}. {\bf{Left to Right:}} For each input reference image, we show its ground-truth depth map (GT Depth), depth map reported by DMDE \cite{ranftl2016dense} and depth map obtained using our approach. Note: Dense GT depth map for KITTI Dataset is taken from DMDE \cite{ranftl2016dense} work. }
\label{fig:comparisonwithKoltun}
\end{figure}

\begin{figure}
  \centering
  \includegraphics[width=0.5\textwidth] {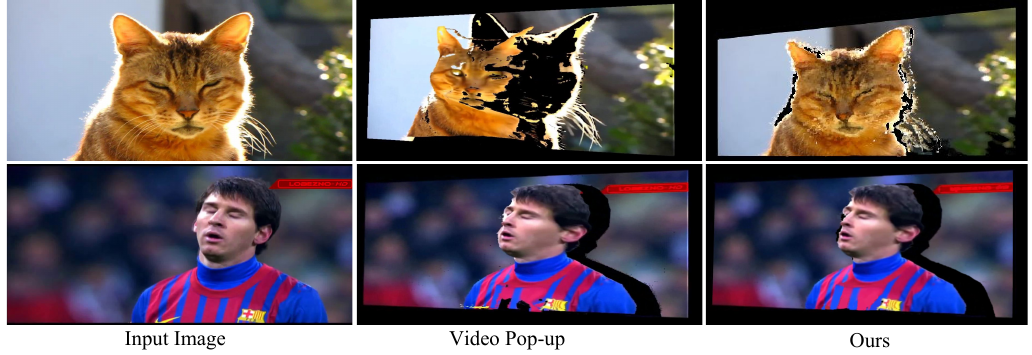}
  \caption{\scriptsize Qualitative evaluation of our approach with the Video-PopUp \cite{russell2014video}. Clearly, our method provides more dense and detailed reconstruction of the scene. In the second row t-shirt description is missing with Video-PopUp \cite{russell2014video} approach. By contrast our method has no such holes. Note: The results presented here for Video-PopUp are taken from their webpage since the source code provided by the authors crashes frequently.}
  \label{fig:videoPopUpComparison}
  \end{figure}

\begin{figure*}
    \begin{center}
    \subfigure[\label{fig:MPIVKITTIStats}]{\includegraphics[width=0.43\linewidth, height=0.17\linewidth ]{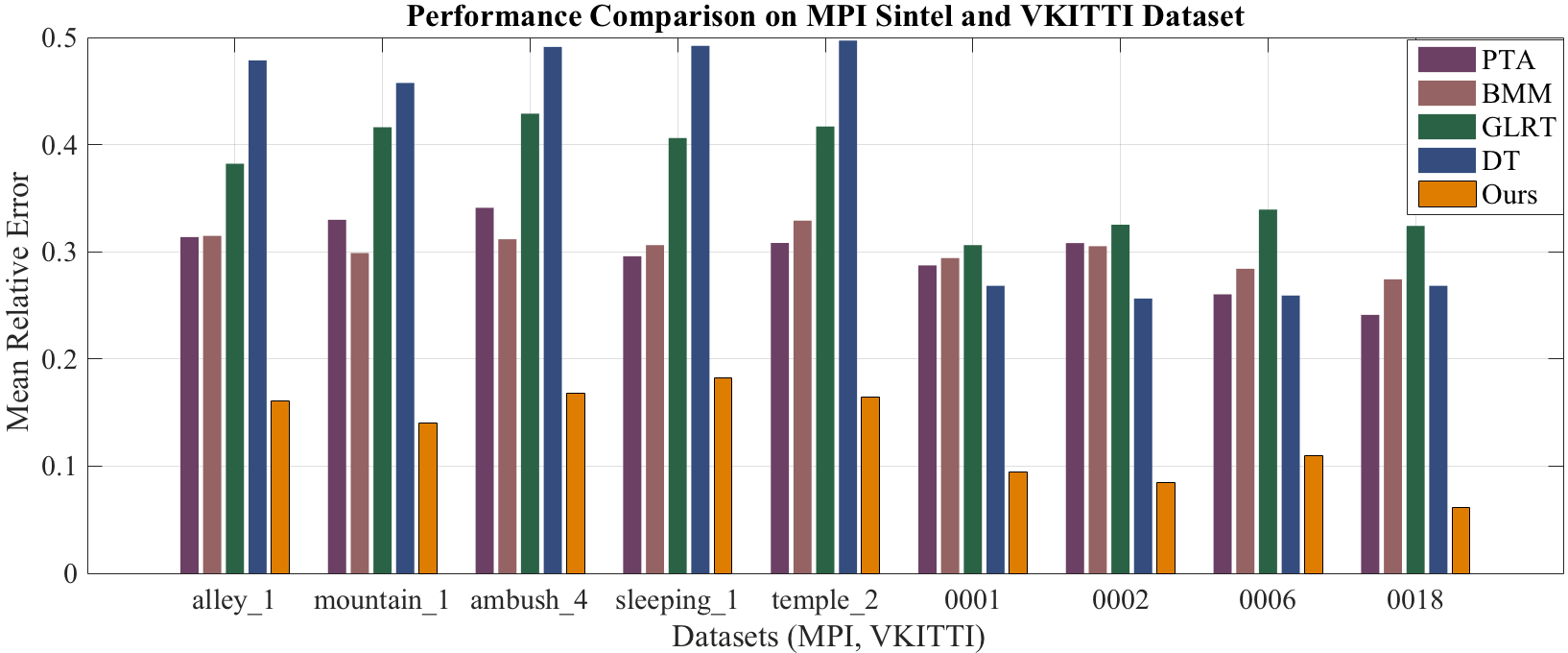}}
    \subfigure[\label{fig:KITTIStats}]{\includegraphics[width=0.43\linewidth, height=0.17\linewidth]{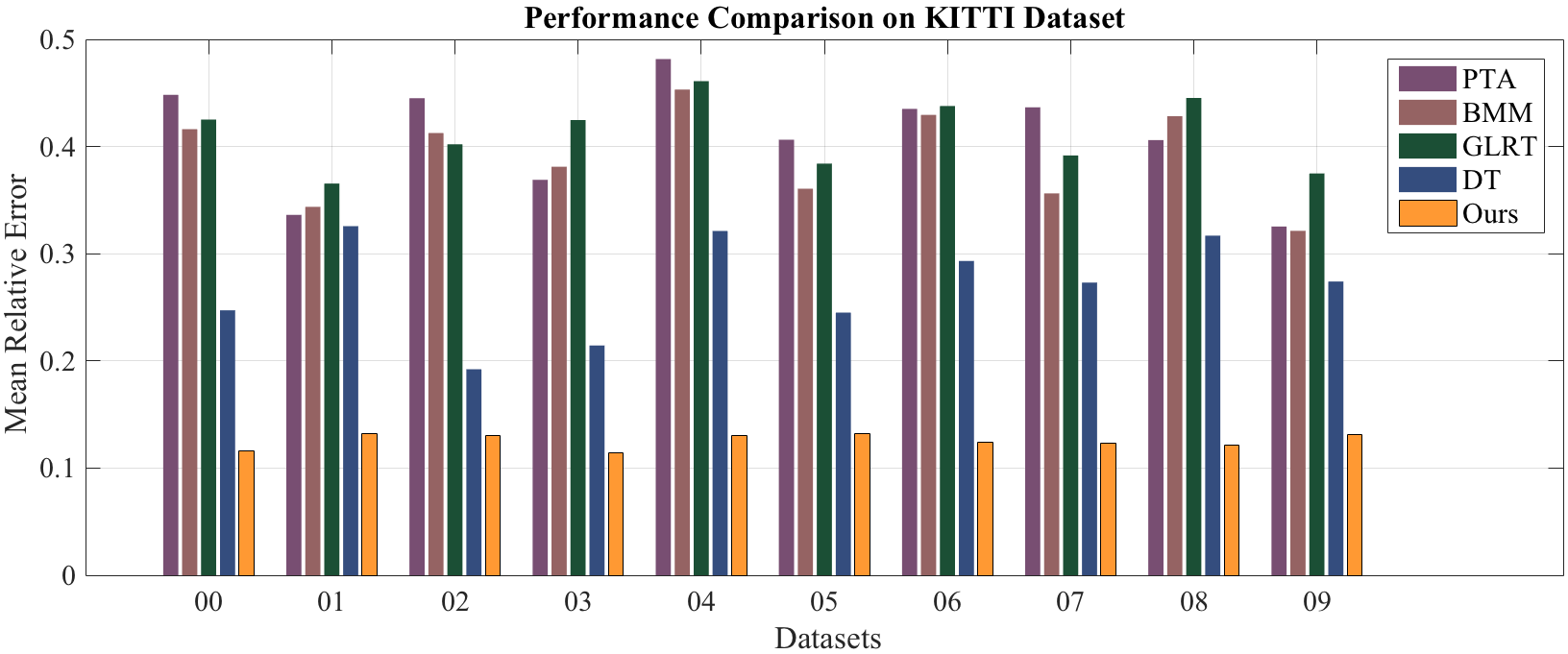}}
  \end{center}
    \caption{\scriptsize Quantitative comparison with our method with  PTA \cite{akhter2011trajectory}, BMM \cite{dai2014simple}, GLRT\cite{fragkiadaki2014grouping}, DT \cite{karsch2014depth} on benchmark datasets. The depth error is calculated by adjusting the numerical scale of the obtained depth map to the ground-truth value, to account for global scale ambiguity. (a)-(b) comparison on MPI Sintel \cite{butler2012naturalistic}, Virtual KITTI \cite{gaidon2016virtual} and KITTI \cite{geiger2013vision} dataset. These numerical values show the fidelity of reconstruction that can be retrieved on these benchmark datasets using our formulation.}
    \label{fig:comparison_vkitti_kitti_mpi}
  \end{figure*}

We compared the performance of our algorithm against several dynamic reconstruction methods, namely, Block Matrix Method (BMM) \cite{dai2014simple}, Point Trajectory 
Approach (PTA) \cite{akhter2011trajectory}, Low-rank Reconstruction (GBLR) \cite{fragkiadaki2014grouping}), Depth Transfer (DT) \cite{karsch2014depth}, DMDE \cite{ranftl2016dense} and ULDEMV \cite{zhou2017unsupervised}. This comparison is made on the available benchmark datasets i.e., MPI Sintel (MPI-S), KITTI, VKITTI, kinect\_tshirt (k\_tshirt), kinect\_paper (k\_paper). Table \ref{tab:comparisonOverall} provides the statistical result of our method in comparison to the baseline approach on these datasets. Our method outperforms others in the outdoor sequence and provides a commendable performance for deformable sequence. Additionally, we performed a qualitative comparison on MPI Sintel \cite{butler2012naturalistic}, KITTI\cite{geiger2013vision} and You-Tube object dataset\cite{prest2012learning}. Fig. \ref{fig:comparisonwithKoltun} and 
Fig. \ref{fig:videoPopUpComparison} provides the visual comparison result of our method to other competing methods. It can be observed that our method consistently delivers superior performance on all of these datasets. While compiling the results per frame comparison is also made over the entire sequence. Evaluation in the case of KITTI dataset is done only for the provided sparse 3D LiDAR points. Fig. \ref{fig:MPIVKITTIStats}, Fig. \ref{fig:KITTIStats} and Fig. \ref{fig:ComparisonwithULDEMV} shows per category statistical performance of our approach against other competing methods on the benchmark dataset.



\subsection{Performance Analysis}
Besides statistical comparison, we conducted other experiments to analyze the behavior of our algorithm. These 
experiments supply an in-depth understanding of the dependency of our algorithm on other input modules.

\begin{table}
\centering
\begin{tabular}{l|l|l|l|l|l|l}
\hline
\begin{tabular}[l]{@{}l@{}}\textbf{Method} \\ Dataset $\downarrow$\end{tabular} & \begin{tabular}[l]{@{}l@{}}DT\\ (SF)\end{tabular} & \begin{tabular}[l]{@{}l@{}}GLRT\\ (MF)\end{tabular} & \begin{tabular}[l]{@{}l@{}}BMM\\ (MF)\end{tabular} & \begin{tabular}[l]{@{}l@{}}PTA\\ (MF)\end{tabular} & \begin{tabular}[l]{@{}l@{}}DMDE \\ (TF)\end{tabular} & \begin{tabular}[c]{@{}l@{}}Ours\\ (TF)\end{tabular} \\ \hline
MPI-S & 0.4833 & 0.4101  & 0.3121 & 0.3177 & 0.297  & {\bf{0.1643}} \\ \hline
V-KITTI & 0.2630 & 0.3237 & 0.2894 & 0.2742 & - & {\bf{0.0925}} \\ \hline
KITTI & 0.2703 & 0.4112 & 0.3903 & 0.4090 & 0.148 & {\bf{0.1254}} \\ \hline
k\_paper & 0.2040 & 0.0920 & {\bf{0.0322}} & 0.0520 & - & 0.0472 \\ \hline
k\_tshirt & 0.2170 & 0.1030 & 0.0443 & {\bf{0.0420}} & - & 0.0480 \\
\hline
\end{tabular}
\caption{\scriptsize Performance Comparison: This table lists the MRE errors. For DMDE \cite{ranftl2016dense} we used its previously reported results as its implementation is not publicly available. SF, MF, TF refers to single frame, multi-frame and two-frame based approach respectively. The reference to the method DT\cite{karsch2014depth}, GLRT\cite{fragkiadaki2014grouping}, BMM\cite{dai2014simple}, PTA\cite{akhter2011trajectory}, DMDE \cite{ranftl2016dense}.}
\label{tab:comparisonOverall}
\end{table}

\begin{figure*}
  \begin{center}
   \subfigure[\label{fig:nsuperpixelVariation}]{\includegraphics[width=0.25\linewidth, height=0.14\linewidth]{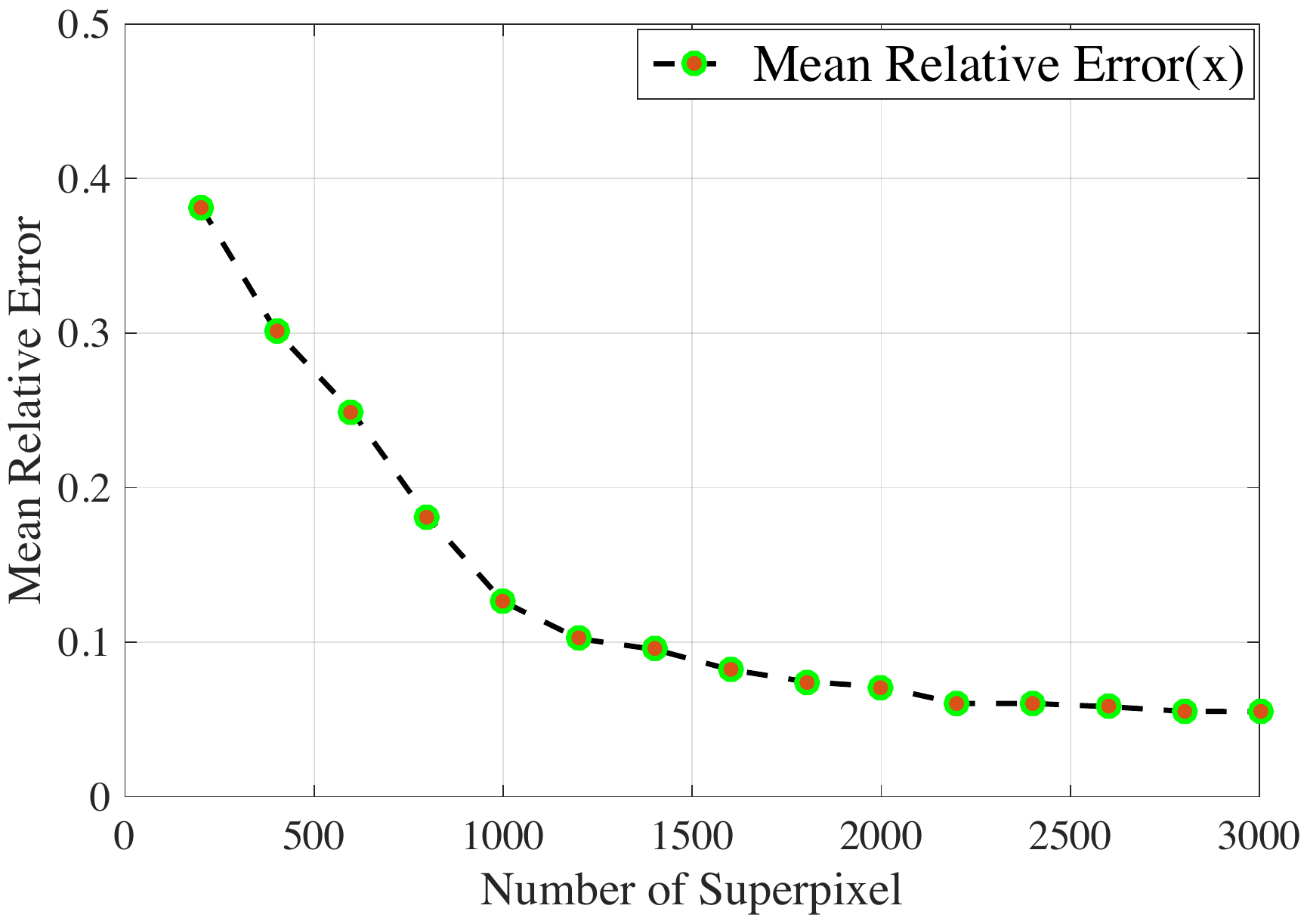}}
    \subfigure[\label{fig:opv}]{\includegraphics[width=0.25\linewidth, height=0.14\linewidth]{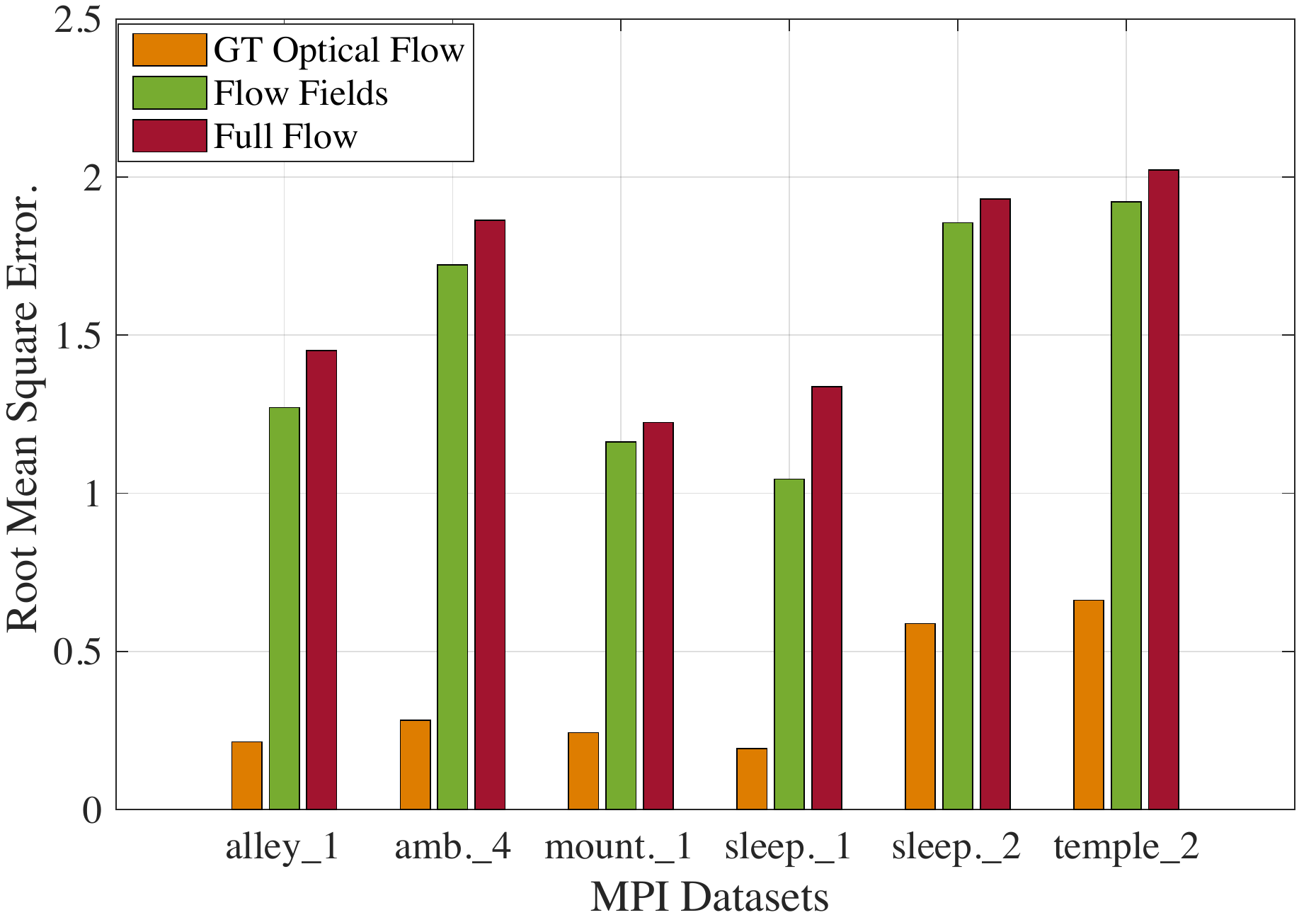}}
    \subfigure[\label{fig:ComparisonwithULDEMV}]{\includegraphics[width=0.25\linewidth, height=0.14\linewidth]{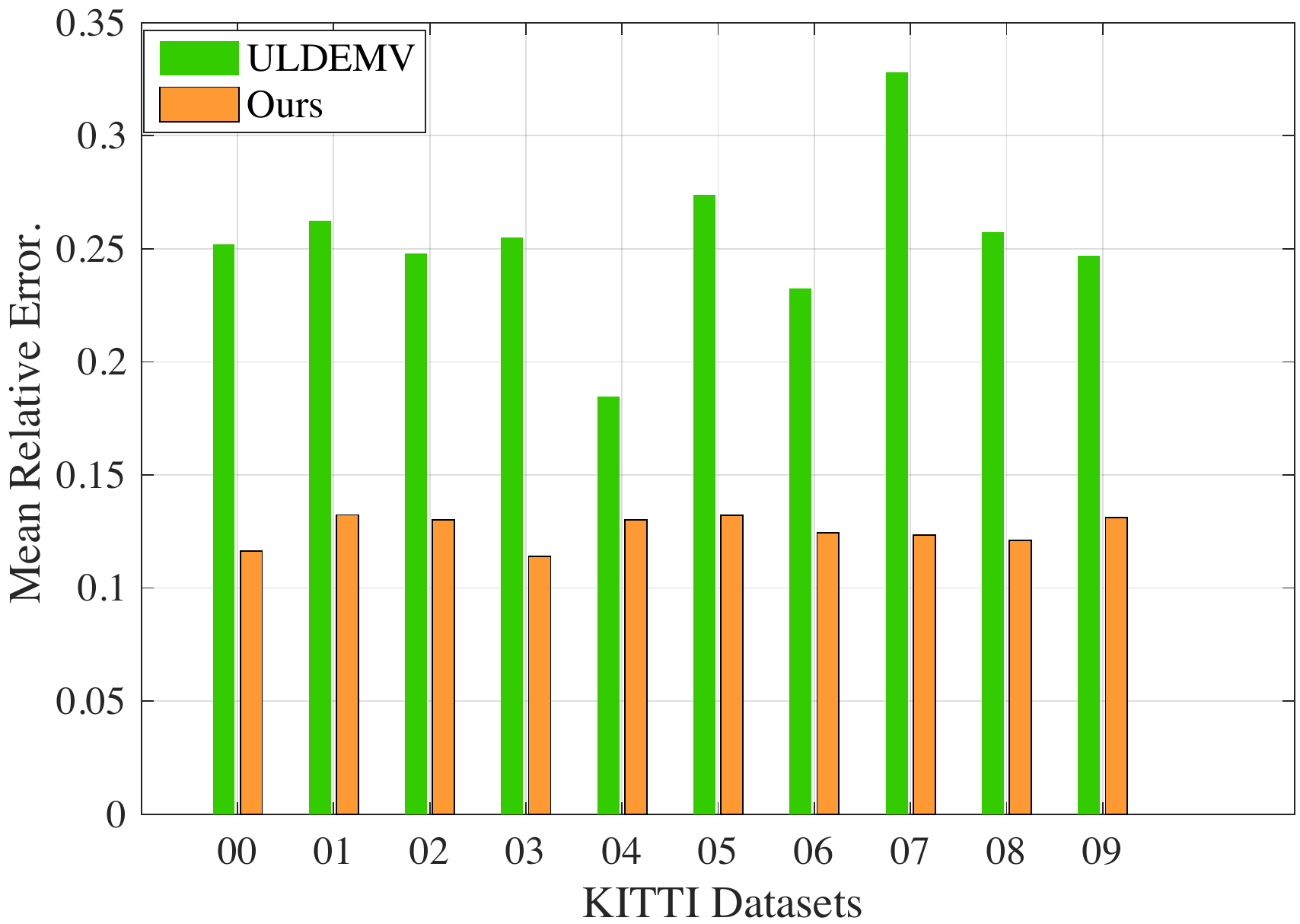}}
  \end{center}
  \caption{ \scriptsize (a) Change in mean relative depth error with the change in number of superpixels. It can be observed that after 1000 superpixel the MRE more or less starts saturating with no significant effect on the  overall accuracy. However, it was observed that the motion estimation becomes critical with the increase in number of superpixels. (b) Performance evaluation in RMSE (in meters) with the state-of-the-art optical flow methods in comparison to the ground-truth optical flow (MPI Sintel \cite{butler2012naturalistic} dataset). (c) Mean Relative Depth Error comparison with a recently proposed unsupervised learning based approach (ULDEMV \cite{zhou2017unsupervised}) on KITTI dataset \cite{geiger2013vision}.}
  \end{figure*}

\textbf{Performance with variation in number of superpixels}:
Our method uses SLIC based over-segmentation of the reference frame to discretize the 3D space. Therefore, the number of superpixels that represent the real-world plays a crucial role in the accuracy of piece-wise continuous reconstruction. If the number of superpixels are very high the estimation of motion parameters becomes tricky and therefore, neighboring superpixels are used to estimate rigid motion which leads to computation challenges. In contrast, small number of superpixels are unable to capture the intrinsic details of a complex dynamic scene. So, a trade-off between the two is often a better choice. Fig. \ref{fig:nsuperpixelVariation} 
shows the plot of depth error variation with the change in the number of superpixels.

\begin{figure}
  \begin{center}
  \includegraphics[width=0.7\linewidth]{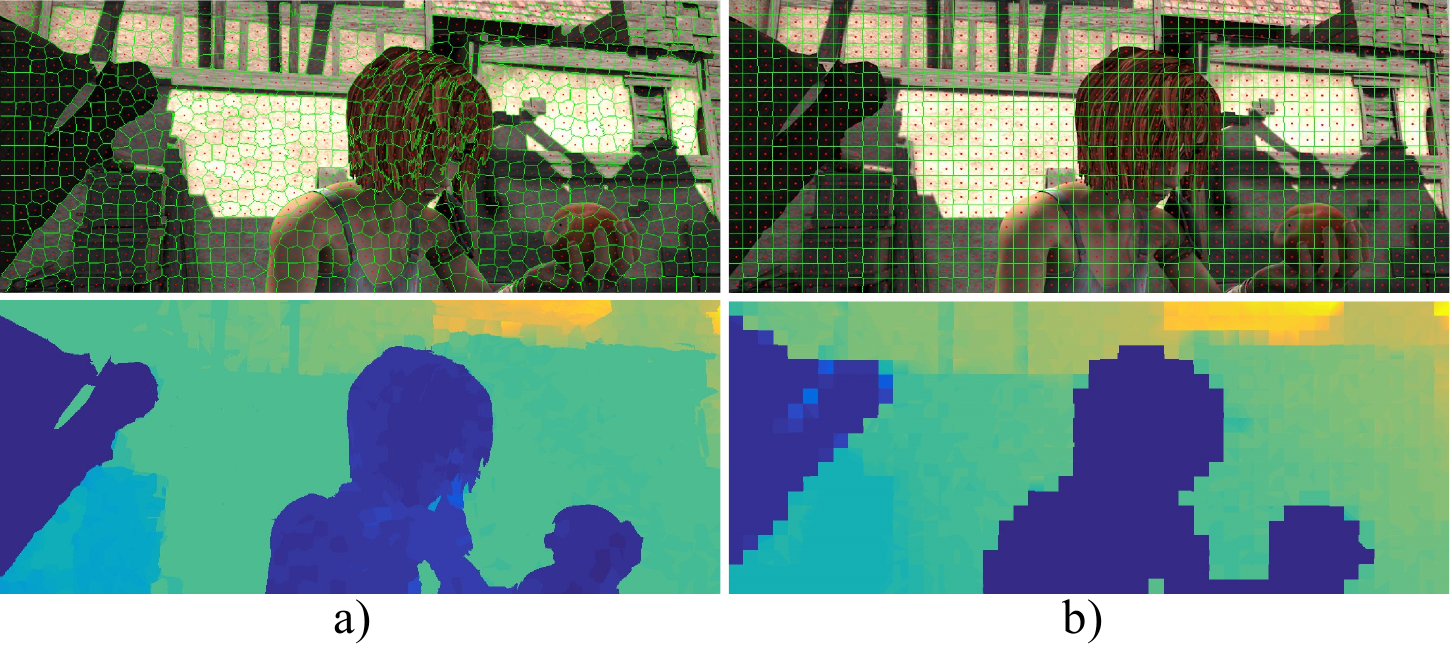}
  \caption{\scriptsize Effects of superpixel pattern on the reconstruction of a dynamic scene. a) with SLIC as superpixels (MRE for the shown frame is 0.0912) b) with uniform grid as superpixels (MRE achieved for the given frame is 0.1442). \label{fig:regularGridComparison}}
  \end{center}
  \end{figure}

  \textbf{Performance with regular grid as image superpixel}:
  Under the piece-wise planar assumption, its not only the number of superpixels that affects the accuracy of reconstruction but also the type of superpixel pattern. To analyze this dependency, we took the worst possible case i.e to divide the reference image into approximately 1000 regular grid and compare its performance against 1000 SLIC superpixel. Our observation clearly shows a decline in the performance in comparison to SLIC superpixels. However, the difference in accuracy is not very significant (see Fig. \ref{fig:regularGridComparison}).
  
\begin{figure}
  \centering
  \includegraphics[width=0.49\textwidth] {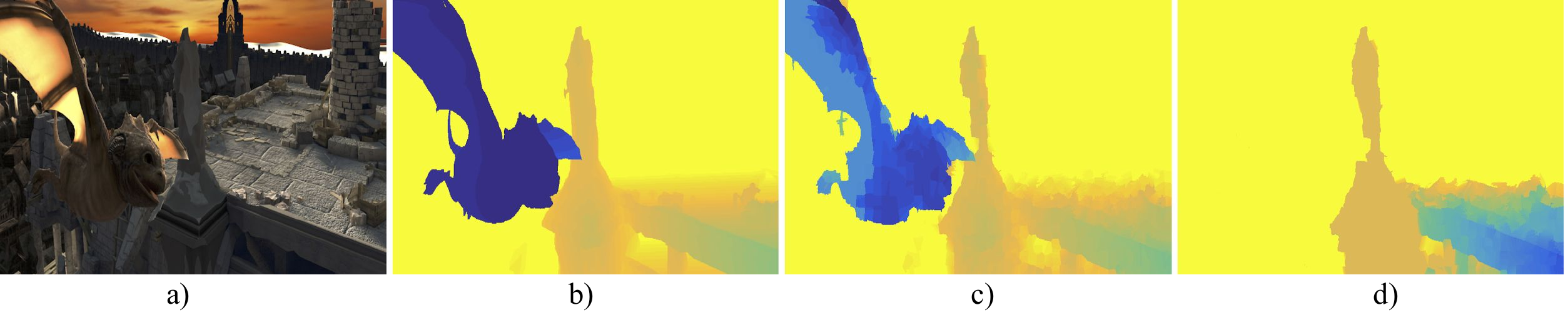}~~~
  \caption{\scriptsize Effect of parameter K in building the K-NN graph.  Our algorithm results in good reconstruction if a suitable K is chosen, in accordance with the levels of complexity in a dynamic scene. (b) Ground-truth depth-map (scaled for illustration purpose). (c) when K=4, a reasonable reconstruction is obtained. (d) when K=20, regions tend to grow bigger. (Best viewed in color.)}
  \label{fig:dragonfailure}
  \end{figure}

  \textbf{Effects of K in K-NN Graph}:
  In our method, the ARAP energy term is evaluated using the K nearest neighbor graph. Different K value leads to different 3D reconstruction result. An experiment on the flying dragon sequence is conducted to analyze the effect of varying K on the performance of our algorithm. The result of the flying dragon case is shown in Fig.~\ref{fig:dragonfailure}. With the increase in K, the rigidity constraint is enforced in an increased neighborhood which directs the 3D reconstruction towards a globally rigid solution. On the other hand, a very small value of K fails to constrain the within object motion. In most of our experiments, we used a K in the range of $15-20$ which achieved satisfactory 3D reconstruction. Also, increasing the value of K  directly affects the overall algorithmic complexity.
  
  \begin{figure}
  \centering
  \includegraphics[width=0.49\textwidth] {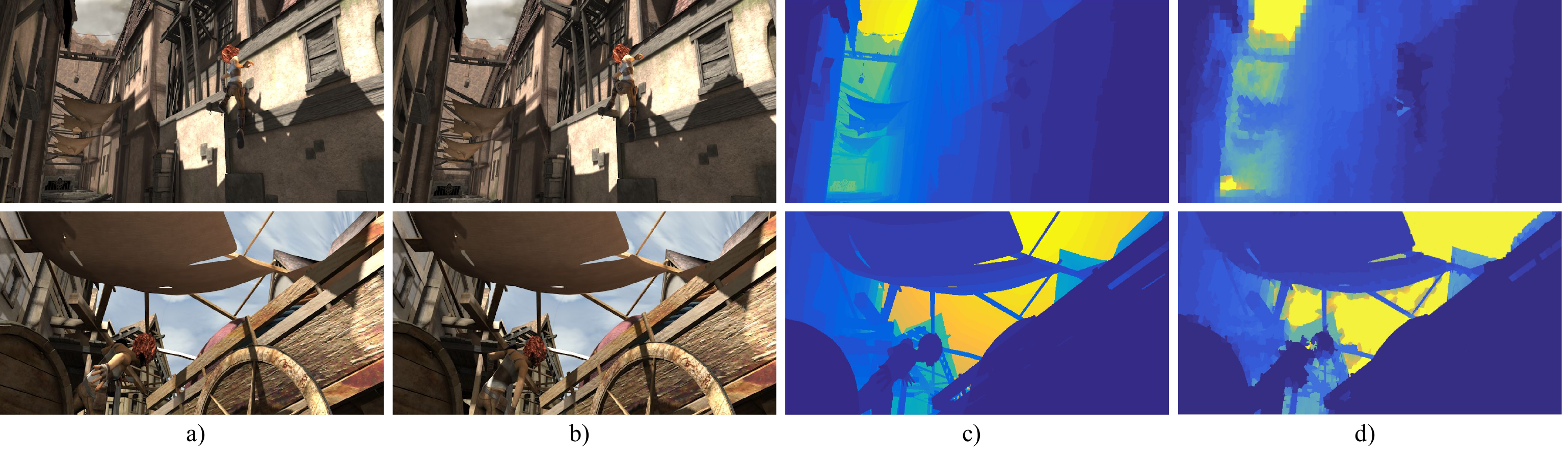}~~~
  \caption{\scriptsize (a)-(b) are the reference frame and the next frame. It is a very challenging case for proper scale recovery with monocular images with dynamic motion. In both of these cases the motion of the girl between two consecutive frames is very large and therefore, the neighboring relations with the planes (say superpixels in image domain) in the consecutive frames gets violated. In such cases, our method may not be able to provide correct scales for each moving planes in 3D. In the first example, the complicated motion of the feet of the girl leads to wrong scale estimation. In the second example, the cart along with girl is  moving w.r.t the camera. The hand of the girl has a substantial motion in the consecutive frames which leads to incorrect estimation of scale. (c)-(d) Ground-truth and obtained depth map respectively.}
  \label{fig:failurecase}
  \end{figure}

\textbf{Performance variation using different optical flow algorithm}:
As our method uses dense optical flow correspondences between frames as input, the performance of our method is directly affected by its accuracy. To analyze the sensitivity of our method to different optical flow methods, we conducted experiments by testing our method with the ground-truth optical flow and few state-of-the-art optical flow methods \cite{bailer2015flow} \cite{chen2016full}. In Fig. \ref{fig:opv}, we show the 3D reconstruction performance evaluated in RMSE \footnote{\scriptsize RMSE (Root Mean Square Error) $ = \sqrt[]{\frac{1}{\m P}\sum_{i=1}^{\m P}(\m z_\textrm{gt}^{\m i} -  z_\textrm{e}^{\m i})^2}~$, here $z_\textrm{e}^{\m i}$,  $\m z_\textrm{gt}^{\m i}$ denotes the estimated and ground-truth depth respectively and $\m P$ is the total number of points.} with different optical flow as inputs. This experiment reveals the importance of dense optical flow in the accurate reconstruction of a dynamic scene. While ground truth optical flow naturally achieves the best performance, the difference in result using different state-of-the-art optical flow is not dramatic. Therefore, we conclude that our method can achieve reliable results with the available dense optical flow algorithm's.

\section{Limitations and Discussion}
The success of our method depends on the effectiveness of the piece-wise planar and as rigid as possible assumption. As a result, our method may fail if the piece-wise smooth model is no longer a valid approximation for the dynamic scene. For example, very fine or very small structures which are considerably far from the camera are difficult to recover under the piecewise planar assumption. Further, what about as rigid as possible assumption, \emph{When as rigid as possible assumption may fail?}
When the motion of the dynamic objects between consecutive frame is significantly large such that most of its neighboring relations in the reference frame get violated in the next frame.  Additionally, if the non-rigid shape shrinks or expands over frames such as a \emph{deflating or inflating balloon}, ARAP model fails. A couple of examples for such situations are discussed in Fig. \ref{fig:failurecase}. The other major limitation of our method is the overall processing time.

\subsection{Discussion}
\emph{1. Direction to reduce the processing time of our algorithm}:
Our algorithm is computationally expensive to execute on a regular desktop machine. This is due to the formulation for solving a higher-order graph optimization problem and particle-based refinement using TRW-S. To speed up the processing time, we are implementing some of the recent research work in the field of fast interior-point optimization and message-passing algorithm \cite{pearson2017fast, Tourani_2018_ECCV} to our framework. We believe solving our optimization using these algorithms along with better computation capabilities can significantly reduce the processing time of our method.

\noindent
\emph{2. Suitability of euclidean distance metric between graph vertices}:
Generally, the euclidean distance metric between graph vertices works well under our piece-wise planar assumption of a dynamic scene. However, there are situations where it may not be an appropriate metric. For example: when the shape of the superpixels is affected by noise or modeling of curved spaces using a piece-wise planar graph structure. To handle such special cases its better to measure distance in an embedding space (isometric embedding) or use $l_1$ metric, etc. To be very precise, depending on the shape of the deforming structure over time, the choice of a suitable metric may vary. Interested readers are encouraged to study the field of intrinsic metric on graphs \cite{keller2015intrinsic}.

\begin{figure}
  \begin{center}
  \includegraphics[width=0.7\linewidth, height=0.4\linewidth]{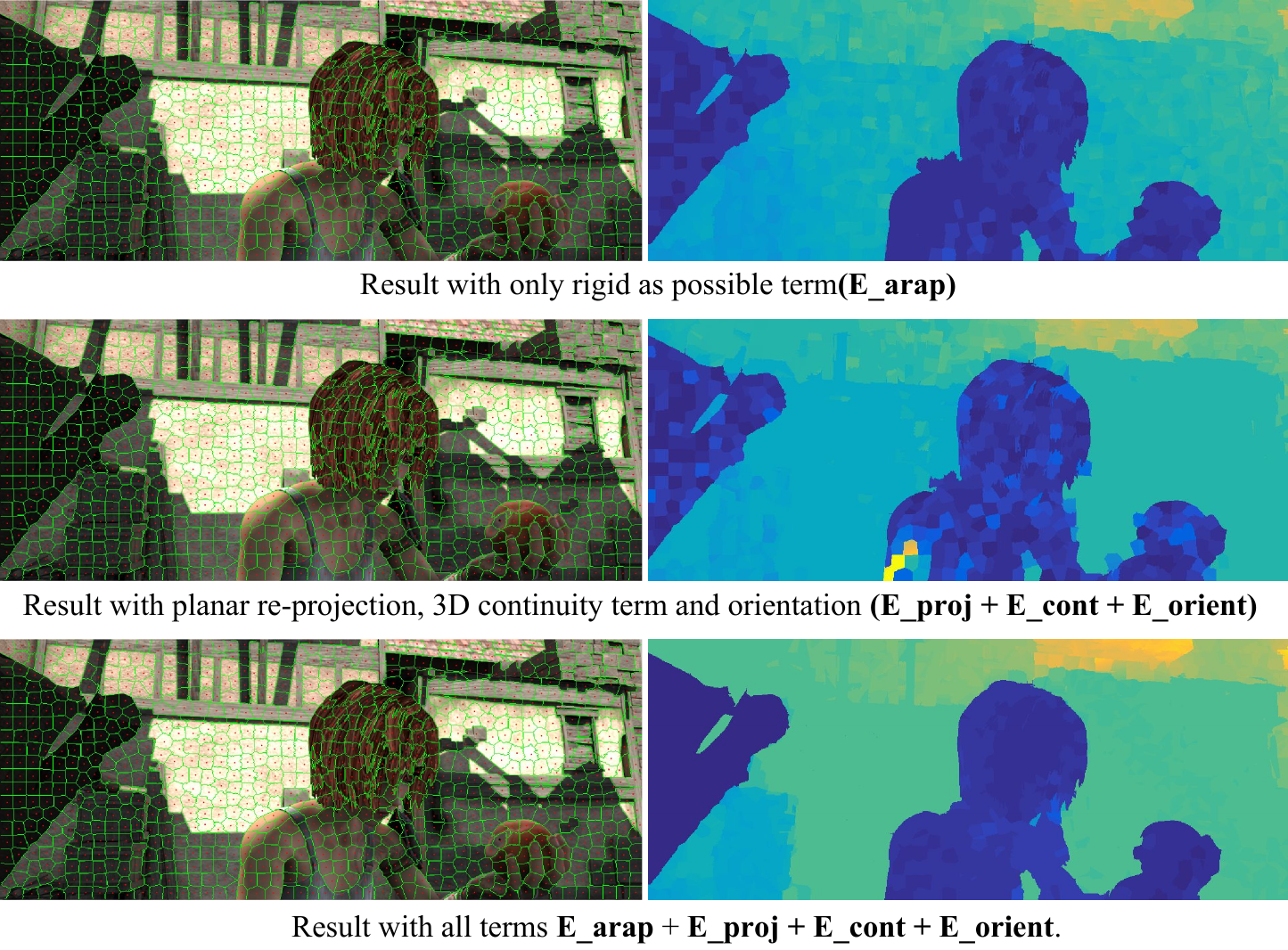}
  \caption{\scriptsize Effect of using ``as rigid as possible'', ``Planar re-projection'', ``3D continuity" and ``Orientation'' term. {\bf{Top row:}} By enforcing the ``as rigid as possible'' term only, the recovered relative scales are correct but the reconstructed planes are misaligned with respect to their neighbors. {\bf{Middle row:} } With the planar re-projection, 3D continuity and orientation term enforced, the resultant 3D reconstruction achieves continuous neighboring boundaries, however, the relative scales for every plane in 3D are not correct.  {\bf{Bottom row:} } By enforcing the the ``as rigid as possible'' term along with all the other smoothness terms, we can handle both relative scales and 3D reconstruction for a complex dynamic scene. \label{fig:eachEnergyTerm}}
  \end{center}
  \end{figure}

\noindent 
\textbf{Ablation Analysis:}
To understand the contribution of different energy term in the overall optimization, we performed ablation analysis.
Firstly, in the proposed optimization framework the {\it{3D continuity}} term is defined over boundaries between neighboring superpixels, which alone is not sufficient to constrain the motion beyond its immediate neighbors.  Secondly, $\m E_\textrm{proj}$ and $\m E_\textrm{orient}$ has nothing to do with scale computation whatsoever. Hence, combining these three terms is not good enough to explain the correct scales for each of the object present in the scene.
On the other hand, {\it{as rigid as possible}} term is defined for each superpixel's anchor point over the K-NN graph structure. However, it does not take into account the alignment of planes in 3D along the boundaries. As a result, the overall reconstruction suffers. Thus, this demonstrates that all the terms are essential for reliable dynamic 3D reconstruction. Fig. \ref{fig:eachEnergyTerm} illustrates the contribution of different terms toward the 
final reconstruction result. Table \ref{tab:ablation_quant} provides numerical value showing the importance of different terms on the overall performance of our algorithm. It can be observed that the improvement in output due to normal orientation constraint is not very significant.

  \begin{table}[h]
  \centering
  \small
  \begin{tabular}{c|c|c|c|c}
  \hline
  Data        & $\m E_\textrm{arap}$  & $+ \m E_\textrm{proj}$  & $+ \m E_\textrm{cont}$ & $+ \m E_\textrm{orient}$ \\ \hline
  alley\_1 &  0.2248   & 0.2022    &  0.1697   & 0.1606 \\ \hline
  ambush\_4 & 0.2381   & 0.2093    &  0.1701   & 0.1676 \\ \hline
  mountain\_1 & 0.2127  & 0.1923    &  0.1492   & 0.1405  \\ \hline
  sleeping\_1 & 0.2418  & 0.2026  &  0.1912   & 0.1823  \\ \hline
  \end{tabular}
  \caption{\scriptsize Contribution of each individual energy term to the overall optimzation. Each column show the mean relative reconstruction error due the addition of the respective energy term. The $+$ sign symbolizes the addition of the all the energy term (columns) left of it.}\label{tab:ablation_quant}
  \end{table}

\section{Conclusion}
In this paper, we have explored, investigated and supplied a distinct perspective to solve one of the classical problems in geometric computer vision i.e., to reconstruct a dense 3D model of a {\em complex, dynamic, and generally non-rigid} scene from its two perspective images. This topic of research is often considered as a very challenging task in structure from motion. In spite of its reasonable challenges, we have demonstrated that dense detailed 3D reconstruction of dynamic scenes is, in fact possible, provided that certain prior assumptions about the scene geometry and the deformation in the scene are satisfied. Both the assumptions we used are mild, realistic and commonly satisfied by the real-world scenarios. Our comprehensive evaluation on the benchmark datasets shows that our new insight to solve dense monocular 3D reconstruction of a general dynamic scene provides better results than other competing methods. This said, we think more profound research on top of our idea may help in the development of sophisticated SfM algorithms. 

\noindent
\textbf{Acknowledgements:} {This research is supported in part by the Australia Research Council  ARC Centre of Excellence for Robotics Vision (CE140100016),  ARC-Discovery (DP 190102261) and ARC-LIEF (190100080), The Natural Science Foundation of China grants (61871325, 61420106007, 61671387), the ``New Generation of Artificial Intelligence'' major project under Grant 2018AAA0102800, and ARC grant DE140100180 and in part by a research gift from Baidu RAL (ApolloScapes-Robotics and Autonomous Driving Lab). The authors gratefully acknowledge the Data Science GPU gift award by NVIDIA Corporation. We thank all the reviewers and AE for their constructive suggestions.}


%





\ifCLASSOPTIONcaptionsoff
  \newpage
\fi



\bibliographystyle{IEEETrans}
\bibliography{final_draft.bib}




%

\begin{IEEEbiography}
[{\includegraphics[width=1in,height=1.25in,clip,keepaspectratio]{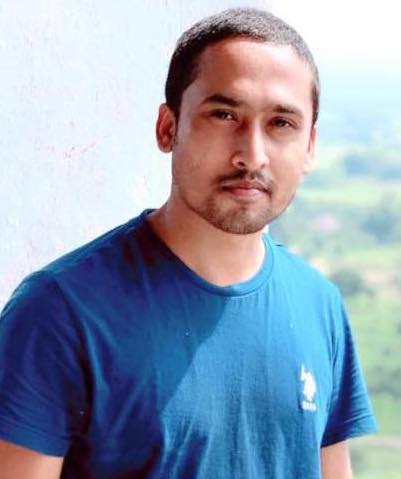}}]
{Suryansh Kumar} is a Professur f\"ur Computer Vision at ETH Z\"urich. He received his Ph.D. in Engineering and Computer Science from the Australian National University, Canberra, Australia. He received his M.S in Computer Science and Engineering from International Institute of Information Technology, Hyderabad (IIIT-H) in 2013. Before joining Australian National University, he worked as a Visiting Scientist in the e-Motion Group at INRIA Rh\^one Alpes Grenoble. After that, he spend one year as a consultant engineer in a computer vision industry-Hyderabad, India. His research interests includes geometric computer vision, robotics, abstract algebra, machine learning and mathematical optimization. He won the best algorithm award in CVPR NRSFM Challenge 2017.
\end{IEEEbiography}

\begin{IEEEbiography}
[{\includegraphics[width=1.0in,height=1.25in,clip,keepaspectratio]{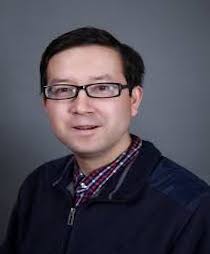}}]
{Yuchao Dai} is a Professor in School of Electronics and Information at Northwestern Polytechnical University, Xi'an, China. He was an ARC DECRA Fellow with the Research School of Engineering at the Australian National University, Canberra, Australia. He received the B.E. degree, M.E degree and Ph.D. degree all in signal and information processing from Northwestern Polytechnical University in 2005, 2008 and 2012, respectively. His research interests include structure from motion, multiview geometry, deep learning, compressive sensing and optimization. He won the Best Paper Award in CVPR 2012, DSTO Best Fundamental Contribution to Image Processing Paper Prize in 2014 and Best Algorithm award in CVPR NRSFM Challenge 2017. He served as an Area Chair for WACV 2019/2020 and ACM MM 2019.
\end{IEEEbiography}


\begin{IEEEbiography}
[{\includegraphics[width=1in,height=1.25in,clip,keepaspectratio]{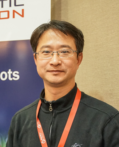}}]
{Hongdong Li}
is a Chief Investigator of the Australia Centre of Excellence for Robotic Vision, Australian National University. He is Associate Director (for Research) for ANU School of Engineering.  He was a visiting professor with the Robotics Institute, CMU during sabbatical in 2017. His research interests include geometric computer vision, pattern recognition and machine learning, vision perception for autonomous driving, and combinatorial optimization.  He is an Associate Editor for IEEE TPAMI, and served as Area Chair for recent years' CVPR, ICCV and ECCV. He was the winner of CVPR Best Paper Award 2012, Marr Prize (Honorable Mention) at ICCV 2017, IEEE ICPR and IEEE ICIP Best Student Paper Winner, DSTO Best Fundamental Contribution to Image Processing Paper Prize at DICTA 2014 and Best algorithm award in CVPR NRSFM Challenge 2017. He is a program co-chair for ACCV 2018 and ACCV 2022.  His research is funded by Australia Research Council, CSIRO, General Motors, Ford Motors, Microsoft Research etc.  He is Presentation Co-Chair for ICCV 2019 and AC for CVPR 2020 and ECCV 2020. \end{IEEEbiography}

\end{document}

%% file: macro3.tex
\newcommand{\SKIP}[1]{} 

\newcommand{\mbegin} {\left [ \begin{array}}

\newcommand{\mend}   {\end{array} \right ]}

\newcommand{\detbegin} {\left | \begin{array}}
\newcommand{\detend}   {\end{array} \right |}

\newcommand{\vbegin} {\left ( \begin{array}{c}}

\newcommand{\vend} {\end{array}\right )}

\def\squareforqed{\hbox{\rlap{$\sqcap$}$\sqcup$}}

\def\qed{\ifmmode\squareforqed\else{\unskip\nobreak\hfil
	\penalty50\hskip1em\null\nobreak\hfil\squareforqed
	\parfillskip=0pt\finalhyphendemerits=0\endgraf}\fi}

\newcommand{\showeqnlabel}{
	\hbox to 0pt{\quad\quad\relax\fbox{\scriptsize\rm\eqnlblx}%
	\gdef\eqnlblx{xxxx}}} \newcommand{\eqnlblx}{}

\def\@eqnnum{\rm (\theequation)\showeqnlabel}

\newcommand{\nofig}[1]{\centerline{\bf Figure here}}

\def\mat#1{\mathchoice{\mbox{\bf$\displaystyle\tt#1$}}
	{\mbox{\bf$\textstyle\tt#1$}}
	{\mbox{\bf$\scriptstyle\tt#1$}}
	{\mbox{\bf$\scriptscriptstyle\tt#1$}}}
\def\m#1{\protect\mat #1}